%% file: main.tex
\documentclass[lettersize,journal]{IEEEtran}
\PassOptionsToPackage{numbers,sort&compress}{natbib}
\usepackage{amsmath,amsfonts}
\usepackage{array}
\usepackage[caption=false,font=normalsize,labelfont=sf,textfont=sf]{subfig}
\usepackage{textcomp}
\usepackage{stfloats}
\usepackage{url}
\usepackage{verbatim}
\usepackage{graphicx}
\usepackage{cite}
\usepackage{color}
\usepackage{tabularx}
\usepackage{multirow}
\usepackage{amsmath}
\usepackage{amssymb}
\usepackage{pifont}
\usepackage{hyperref}
\usepackage{colortbl}
\hypersetup{colorlinks=true,
linkcolor=blue,
anchorcolor=blue,
citecolor=blue}
\newcolumntype{C}[1]{>{\centering\arraybackslash}p{#1}}
\usepackage{bm}
\usepackage{algorithm}
\usepackage[noend]{algpseudocode}
\usepackage[table,dvipsnames]{xcolor}
\definecolor{bestcolor}{gray}{.9}
\newcommand{\bestcell}[1]{\cellcolor{bestcolor}{#1}}
\newcommand{\norm}[1]{\left\lVert#1\right\rVert}
\usepackage{setspace}

\begin{document}

\title{StarPose: 3D Human Pose Estimation via Spatial-Temporal Autoregressive Diffusion}

\author{Haoxin~Yang,
        Weihong~Chen,
        Xuemiao~Xu,~\IEEEmembership{Member,~IEEE,}
        Cheng~Xu,
        Peng~Xiao, \\
        Cuifeng~Sun,
        Shaoyu~Huang,
        and Shengfeng He,~\IEEEmembership{Senior Member,~IEEE}
\thanks{This work is supported by Guangdong Provincial Natural Science Foundation for Outstanding Youth Team Project (No. 2024B1515040010), China National Key R\&D Program (Grant No. 2023YFE0202700, 2024YFB4709200), Key-Area Research and Development Program of Guangzhou City (No.2023B01J0022), NSFC Key Project (No. U23A20391), the AI Singapore Programme under the National Research Foundation Singapore (Grant AISG3-GV-1623-011), and the Lee Kong Chian Fellowships. \textit{(Haoxin~Yang and Weihong~Chen contributed equally to this work.) (Corresponding authors: Xuemiao~Xu; Cheng~Xu.)}}
\thanks{Haoxin~Yang, Weihong~Chen, Xuemiao~Xu and Peng~Xiao are with the School of Computer Science and Engineering, South China University of Technology, Guangzhou, China. Xuemiao Xu is also with the State Key Laboratory of Subtropical Building Science, Ministry of Education Key Laboratory of Big Data and Intelligent Robot, and Guangdong Provincial Key Lab of Computational Intelligence and Cyberspace Information, Guangzhou 510640, China. E-mail: harxis@outlook.com, cswilliam@mail.scut.edu.cn; xuemx@scut.edu.cn; cs\_xiaopeng@mail.scut.edu.cn; }
\thanks{Cheng~Xu is with the Centre for Smart Health, The Hong Kong Polytechnic University, Hong Kong. E-mail: cschengxu@gmail.com.}
\thanks{Cuifeng~Sun is with the Cloud Computing Center, Chinese Academy of Sciences, Dongguan 523808, China. E-mail: zhengliling@casc.ac.cn.
}
\thanks{Shaoyu~Huang is with the Guangzhou Yichuang Information Technology Co., Ltd., Guangzhou 510640, China. E-mail: winsy@163.com.
}
\thanks{Shengfeng He is with the School of Computing and Information Systems, Singapore Management University, Singapore. E-mail: shengfenghe@smu.edu.sg.}}

\markboth{IEEE TRANSACTIONS ON CIRCUITS AND SYSTEMS FOR VIDEO TECHNOLOGY}%
{Yang \MakeLowercase{\textit{et al.}}: StarPose: 3D Human Pose Estimation via Spatial-Temporal Autoregressive Diffusion}

\maketitle

\input{sec/abs}
\input{sec/intro}
\input{sec/relatedwork}
\input{sec/method}

\input{sec/exp}
\input{sec/conclusion}

\bibliographystyle{IEEEtran}
\bibliography{My_ref}

\input{sec/bio.tex}

\vfill

\end{document}

%% file: sec/abs.tex
\begin{abstract}
Monocular 3D human pose estimation remains a challenging task due to inherent depth ambiguities and occlusions. Compared to traditional methods based on Transformers or Convolutional Neural Networks (CNNs), recent diffusion-based approaches have shown superior performance, leveraging their probabilistic nature and high-fidelity generation capabilities. However, these methods often fail to account for the spatial and temporal correlations across predicted frames, resulting in limited temporal consistency and inferior accuracy in predicted 3D pose sequences. To address these shortcomings, this paper proposes \textit{StarPose}, an autoregressive diffusion framework that effectively incorporates historical 3D pose predictions and spatial-temporal physical guidance to significantly enhance both the accuracy and temporal coherence of pose predictions. Unlike existing approaches, \textit{StarPose} models the 2D-to-3D pose mapping as an autoregressive diffusion process. By synergically integrating previously predicted 3D poses with 2D pose inputs via a Historical Pose Integration Module (HPIM), the framework generates rich and informative historical pose embeddings that guide subsequent denoising steps, ensuring temporally consistent predictions. In addition, a fully plug-and-play Spatial-Temporal Physical Guidance (STPG) mechanism is tailored to refine the denoising process in an iterative manner, which further enforces spatial anatomical plausibility and temporal motion dynamics, rendering robust and realistic pose estimates. Extensive experiments on benchmark datasets demonstrate that \textit{StarPose} outperforms state-of-the-art methods, achieving superior accuracy and temporal consistency in 3D human pose estimation. Code is available at \url{https://github.com/wileychan/StarPose}.

\end{abstract}


%% file: sec/intro.tex
\section{Introduction}
\IEEEPARstart{3}{D} human pose estimation (HPE) aims to predict the three-dimensional coordinates of human joints using monocular 2D video inputs. As a foundational preprocessing step, it enables a wide range of downstream applications, such as action recognition~\cite{Rajasegaran2023cvpr, lin2024mutual, mao2023masked}, autonomous driving~\cite{Occ3DNEURIPS2023, zanfir23a, zheng2022multi}, and motion capture~\cite{Athanasiou_2023_ICCV, yu2023Toward, xiaopeng_eccv}. The task typically involves two key stages: first, estimating 2D joint positions using established 2D keypoint detectors~\cite{chen2018cascaded}, and second, mapping these 2D keypoints to 3D human poses. This work focuses on the latter stage, commonly referred to as the 2D-to-3D lifting process~\cite{Xu_2024_CVPR, Peng_2024_CVPR, eccv_repose, Gong_2023_CVPR, D3DP2023_ICCV, Tang_2023_CVPR}. 

\begin{figure}
\centerline{\includegraphics[width=0.9\columnwidth]{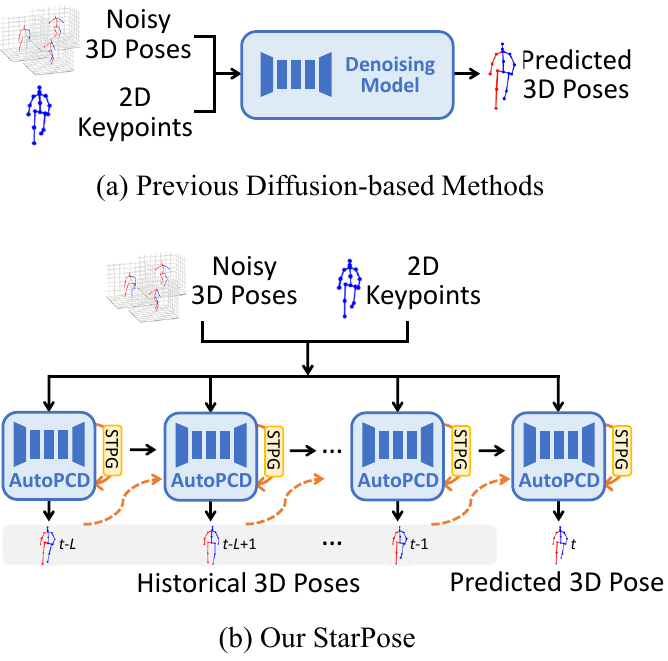}} 
    \caption{(a): Existing diffusion-based methods for 3D HPE~\cite{Xu_2024_CVPR, Gong_2023_CVPR, D3DP2023_ICCV} predict single-frame 3D poses using neighboring 2D poses as conditional inputs within a standard diffusion framework, which are prone to inaccuracies and lacking temporal coherence in predicted 3D pose sequences due to depth ambiguity. To overcome this, we introduce (b) a spatial-temporal autoregressive diffusion framework that strategically leverages historically predicted 3D pose predictions and incorporates spatial-temporal physical constraints during denoising, enabling accurate and temporally consistent 3D pose estimation.
    }
    \label{fig:teaser}
\end{figure}

In the realm of 3D HPE, early techniques leveraged convolutional neural networks or autoregressive long short-term memory (LSTM) networks to achieve pose predictions \cite{zhang2023learning, du2024anatpose, du2023bi, qiu2023weakly, du2024joypose}. More recent advancements~\cite{eccv_repose, Tang_2023_CVPR, Mehraban_2024_WACV, zheng20213D, chen2025scjd} have adopted transformer architectures~\cite{vaswani2017attention} to capture spatial and temporal dependencies. While effective at modeling complex relationships, these methods rely on deterministic regression that maps 2D pose sequences to a single 3D configuration. This formulation neglects the inherent ambiguity of lifting 2D poses to 3D, where a single 2D input can correspond to multiple valid 3D interpretations. Consequently, the above-mentioned methods often produce overconfident and inaccurate results, particularly in cases of occlusion or self-contact, such as crossed limbs.
Diffusion-based approaches~\cite{Xu_2024_CVPR, Gong_2023_CVPR, D3DP2023_ICCV, shan2024diffusion, chen2024diffusion} offer a promising alternative. By drawing on the stochastic sampling process of diffusion models~\cite{ddpmNEURIPS2020}, these methods generate diverse 3D pose hypotheses conditioned on 2D inputs. This process inherently supports multiple plausible outputs and enhances robustness to uncertainty. 
Despite these advantages, existing diffusion-based methods generally treat each 2D pose sequence in isolation, failing to utilize the temporal continuity of human motion. By conditioning only on 2D observations and ignoring prior 3D predictions, they overlook the sequential consistency and causality of motion, which results in temporally unstable and physically implausible predictions (see Fig.~\ref{fig:teaser}(a)).

Human motion unfolds continuously, where each pose intrinsically influences subsequent ones. In the context of 3D HPE, this implies that each predicted pose should be consistent not only with the current 2D observation but also with the sequence of preceding 3D predictions. However, existing diffusion-based approaches~\cite{Xu_2024_CVPR, Gong_2023_CVPR, D3DP2023_ICCV, chen2024diffusion} overlook this fundamental aspect, treating each frame in isolation and ignoring the dynamic nature of human motion.
In addition to temporal coherence, human movement adheres to rigid skeletal structures and biomechanical constraints that encode strong spatial and temporal priors. Embedding these constraints into the modeling process significantly enhances the accuracy and stability of diffusion-based estimators. By enforcing anatomical plausibility and physical consistency, such priors help avoid unrealistic joint configurations and mitigate the accumulation of prediction errors over time. Leveraging this domain knowledge is essential for generating realistic and physically plausible motion trajectories.

Building upon the above observations, we introduce \textbf{StarPose}, a novel \textbf{S}patial-\textbf{T}emporal \textbf{A}uto\textbf{R}egressive diffusion framework designed to achieve precise and temporally consistent 3D HPE.
Unlike conventional diffusion-based methods~\cite{Xu_2024_CVPR, Gong_2023_CVPR, D3DP2023_ICCV}, which generate 3D poses independently for each frame, StarPose employs an autoregressive diffusion mechanism. By iteratively incorporating previously predicted 3D poses and 2D pose conditions as inputs to the denoising model, StarPose effectively leverages historical cues to align with current observations. This design ensures enhanced spatial and temporal consistency in the resulting 3D pose sequences.
As depicted in Fig.~\ref{fig:teaser} (b), StarPose consists of two key components: \textit{Autoregressive Pose Conditional Diffusion} (AutoPCD) and \textit{Spatial-Temporal Physical Guidance} (STPG). AutoPCD captures intricate context dependencies between historical and future poses, by reintroducing previously predicted 3D poses and 2D pose conditions into the model through an autoregressive mechanism. This process is facilitated by the \textit{Historical Pose Integration Module} (HPIM), which fuses ground-truth 2D poses with prior 3D predictions to ensure continuity, reduce drift, and mitigate cumulative error across time.
On the other hand, SPTG incorporates structural and kinematic constraints inherent to human motion and anatomy without the need for ground truth poses. Applied during both training and inference stages, this guidance further enforces spatial and temporal consistency through four key elements: \textit{2D Reprojection Consistency}, \textit{Skeletal Symmetry Penalty}, \textit{Bone Length Variance}, and \textit{Differential Sequence Variation}. These elements collectively promote anatomical plausibility, preserve motion continuity, and further mitigate long-term prediction errors.
Comprehensive evaluations on the Human3.6M~\cite{ionescu2013human3} and MPI-INF-3DHP~\cite{mehta2017monocular} datasets demonstrate the effectiveness of StarPose. Both quantitative and qualitative results confirm its superiority over state-of-the-art methods, achieving accurate and temporally consistent 3D HPE.

Our contributions can be summarized as follows:
\begin{itemize}
    \item We propose StarPose, an innovative Spatial-Temporal AutoRegressive diffusion framework for 3D HPE. By introducing an autoregressive denoising diffusion process, StarPose effectively captures long-term dependencies from previous predictions to enhance future estimates, with explicit human physical guidance from spatial and temporal domains, achieving precise 3D pose estimation with superior temporal coherence.

    \item We develop an autoregressive pose conditional diffusion model that fully exploits the synergy between historically predicted 3D skeletons and corresponding 2D skeletons. This model produces rich, informative historical spatial and temporal cues for the diffusion process, ensuring consistent and accurate predictions across frames.

    \item We explicitly incorporate a comprehensive set of physical priors into the diffusion process by presenting a spatial-temporal physical guidance mechanism. It enforces constraints on both spatial topology and temporal dynamics, leading to anatomically plausible and temporally coherent pose estimations during both training and inference.

    \item Extensive experiments on two widely-used benchmarks demonstrate the superiority of our method over state-of-the-art approaches in delivering accurate and visually compelling 3D human poses from monocular 2D observations.
    
\end{itemize}

%% file: sec/relatedwork.tex
\section{Related Works}
\subsection{Transformer-based 3D Human Pose Estimation}
2D-to-3D lifting methods aim to estimate the corresponding 3D human poses from 2D pose estimates, which are typically derived from images or video frames.
Traditional methods use convolutional neural networks or long short-term memory (LSTM) networks to achieve 3D pose estimation~\cite{zhang2023learning, du2024anatpose,du2023bi,qiu2023weakly,du2024joypose}.
In recent years, with the growing success of vision transformers~\cite{dosovitskiy2021image} in various vision tasks, transformer-based approaches for 3D HPE have gained significant attention. Several studies~\cite{Peng_2024_CVPR, eccv_repose, Tang_2023_CVPR, Mehraban_2024_WACV, ijcai23hdformer, tang2023ftcm, zhang2022mixste, shan2022p, li2022mhformer, zhang2022uncertainty, li2022exploiting, zheng20213D, shan2021improving, chen2021anatomy} have explored the potential of transformers in this domain. 
Most recent methods leverage multi-frame input sequences to capture temporal dependencies, with PoseFormer~\cite{zheng20213D} being one of the first to model both spatial and temporal information for 3D pose estimation. MixSTE~\cite{zhang2022mixste} employs a stacked architecture of spatial and temporal transformer blocks to capture spatial-temporal features alternately, effectively modeling the joint trajectory over a sequence of frames. STCFormer~\cite{Tang_2023_CVPR} slices the input joint features into two partitions and uses self-attention to model spatial and temporal contexts in parallel. 
While these transformer-based methods successfully model spatial and temporal dependencies through attention mechanisms, they are inherently limited by the depth ambiguity that exists in the mapping from 2D skeletons to 3D poses—where a single 2D pose may correspond to multiple 3D solutions. As a result, these methods typically predict only one 3D pose solution for a given 2D input, leading to potential inaccuracies in depth estimation.

\subsection{Diffusion-based 3D Human Pose Estimation}
Diffusion models have recently emerged as a promising paradigm in HPE~\cite{feng2023diffpose}, particularly in the context of 3D pose estimation~\cite{Xu_2024_CVPR, Gong_2023_CVPR, D3DP2023_ICCV, Holmquist_2023_ICCV, choi2023diffupose, Zhou2023diff3dhpe, chen2024diffusion, shan2024diffusion}. These approaches typically formulate 3D pose estimation as a distribution-to-distribution transformation, where the reverse diffusion process iteratively refines an uncertain 3D pose distribution into a more deterministic one, conditioned on input 2D pose sequences~\cite{Gong_2023_CVPR}.
Recent studies~\cite{Xu_2024_CVPR, Gong_2023_CVPR, D3DP2023_ICCV, Holmquist_2023_ICCV, choi2023diffupose, Zhou2023diff3dhpe,chen2024diffusion,shan2024diffusion} have explored the use of diffusion models to address depth ambiguity by treating it as noise and enhancing 2D-to-3D lifting results. In these methods, 2D information serves as a condition for the denoising process. For instance, Holmquist \textit{et al.}~\cite{Holmquist_2023_ICCV} use a transformer-based model to encode the 2D heatmap, which is then passed as a condition to guide the denoising process for 3D pose estimation. Similarly, FinePose~\cite{Xu_2024_CVPR} takes a multi-granular approach, encoding each part of the human body separately to generate more detailed body part-specific information, which is used as conditions to refine the pose prediction.
Moreover, D3DP~\cite{D3DP2023_ICCV, shan2024diffusion} select the most plausible 3D hypothesis by reprojecting the generated 3D poses to the 2D camera plane and choosing the hypothesis with the smallest reprojection error. In these works~\cite{Holmquist_2023_ICCV, Xu_2024_CVPR, D3DP2023_ICCV}, the reverse diffusion process generally begins with random noise, which is gradually denoised to generate an output 3D pose.
Differently, DiffPose~\cite{Gong_2023_CVPR} acknowledges the unique uncertainty of human pose in 3D space and proposes that the denoising process should not start from random noise but rather from a distribution informed by sample-specific knowledge, which initializes the 3D pose distribution according to the input 2D pose. While recent diffusion-based approaches~\cite{Xu_2024_CVPR, Gong_2023_CVPR, D3DP2023_ICCV, Holmquist_2023_ICCV} have demonstrated promising results by conditioning on 2D poses, they generally treat each denoising step in isolation, without incorporating explicit structural constraints between the predicted 3D pose and prior knowledge of human anatomy. These methods typically generate 3D poses independently for each frame, thereby disregarding temporal dependencies and resulting in suboptimal motion continuity and temporal coherence.
In contrast, we introduce a spatial-temporal autoregressive pose-conditional diffusion framework with physical guidance, which fundamentally differs from vanilla diffusion models. Our approach leverages an autoregressive formulation that recursively integrates previously predicted 3D poses into the denoising trajectory. This autoregressive mechanism enables the model to propagate temporal context across frames while jointly enforcing spatial and temporal physical constraints at each step. As a result, our framework yields more accurate and temporally consistent 3D pose estimations, capturing both instantaneous anatomical validity and long-range motion coherence.

\subsection{Autoregressive Diffusion Models}
Diffusion models have shown remarkable capabilities in generation tasks~\cite{DisenDreamer-tcsvt24, 10633292, jingjing2023tcsvt}, and recently, their evolution into autoregressive diffusion models has achieved remarkable progress in image and video generation.
The visual autoregressive model (VAR) model~\cite{tian2024visual} introduces an innovative next-scale prediction paradigm by utilizing a multiscale VQVAE architecture combined with a VAR transformer structure. Meanwhile, another line of work~\cite{sun2024autoregressive} applies the next-token prediction paradigm, where models such as Llama, when appropriately scaled, deliver state-of-the-art performance in image generation. Additionally, a certain approach~\cite{xie2024progressive} assigns progressively increasing noise levels to latent frames instead of relying on a single noise level, thereby enabling the generation of long videos while preserving inter-frame coherence. Another model incorporates GPT-style autoregressive generation into video diffusion frameworks~\cite{gao2024vid}, leveraging causal generation, frame-as-prompt mechanisms, and KV-cache techniques, resulting in outstanding long-video generation capabilities. HPDM~\cite{skorokhodov2024hierarchical} optimizes it for efficient training and end-to-end high-resolution video generation.
Different from existing endeavors, this paper makes the first attempt to introduce a spatial-temporal autoregressive diffusion framework for 3D pose estimation by incorporating historical predictions and physical typology and dynamics guidance during the denoising process, enabling accurate and reliable pose predictions.

%% file: sec/method.tex
\section{Proposed Method}
\begin{figure*}
    \centering
    \centerline{\includegraphics[width=0.9\linewidth]{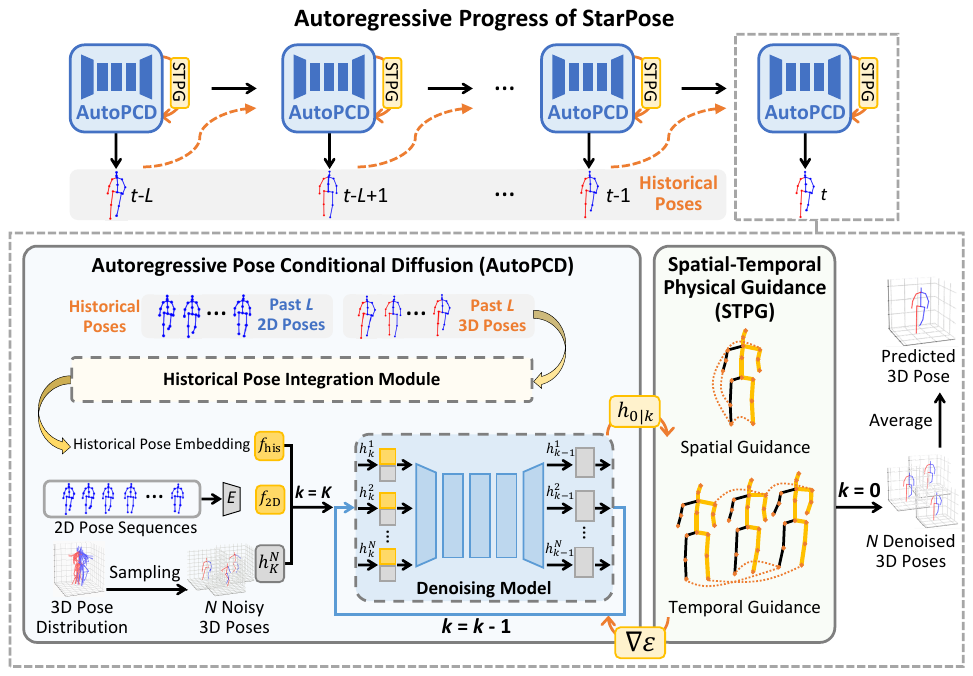}}
    \caption{Overview of the proposed StarPose, which comprises an \textit{Autoregressive Pose Conditional Diffusion} (AutoPCD) and a \textit{Spatial-Temporal Physical Guidance} (STPG). The AutoPCD autoregressively leverages historically predicted 3D poses as conditional input for the current denoising iteration, which is consolidated by the \textit{Historical Pose Integration Module} (HPIM) to ensure consistency in cross-frame pose prediction.
    During the denoising process, we begin by sampling $N$ noisy poses, which are subsequently refined over $K$ iterations. In each iteration, the STPG explicitly regulates the optimization through both spatial and temporal physical constraints on human dynamics, thereby enhancing the accuracy and vividness of pose predictions. The final 3D pose is derived by averaging the $N$ denoised poses.
    }
    \label{fig:framework}
    \vspace{-5pt}
\end{figure*}

The overview of our framework \textbf{StarPose} is shown in Fig.~\ref{fig:framework}, which mainly consists of two components: an \textit{Autoregressive Pose Conditional Diffusion} (AutoPCD) that performs the diffusion steps in the autoregressive process, and a \textit{Spatial-Temporal Physical Guidance} (STPG) which incorporates structural and kinematic constraints inherent to human motion and anatomy to improve the accuracy of 3D HPE.
For convenience, commonly used symbols are summarized in Table~\ref{para}.
\input{table/params}

\subsection{Preliminary for Diffusion-based 3D HPE}
Given the input 2D pose sequence \(X \in \mathbb{R}^{f \times (J \cdot 2)}\) of \(f\) frames, the goal of 3D HPE is to accurately predict the 3D coordinates of $J$ keypoints for the central frame of the sequence, represented as \(y \in \mathbb{R}^{1 \times (J \cdot 3)}\). Diffusion-based generative models can recurrently shed the indeterminacy in an initial distribution to reconstruct a high-quality determinate sample, making them well-suited for addressing the inherent uncertainty and ambiguity in 3D pose estimation. In diffusion models, the \textit{forward} diffusion process gradually adds noise to a clean sample $h_0$, producing a highly indeterminate sample $h_K$ through a series of stochastic steps. This mimics the degradation of information. The \textit{reverse} diffusion process, also known as denoising, aims to reconstruct the original clean sample $h_0$ from the noisy sample $h_K$ through iterative denoising steps. In the context of 3D HPE, the task is framed as constructing a determinate 3D pose distribution (\( H_0 \)) from a highly indeterminate pose initial distribution (\( H_K \)) using diffusion models.

\textbf{Training Process.} 
While \( H_K \) represents a noisy pose distribution with potential errors caused by depth ambiguities, it is distinct from Gaussian white noise. Thus, similar to DiffPose~\cite{Gong_2023_CVPR}, we adopt a Gaussian Mixture Model (GMM) to model this noisy distribution $H_K$. 
Following~\cite{Gong_2023_CVPR}, the GMM is fitted on the training set using the Expectation-Maximization (EM) algorithm to approximate the 2D and 3D distributions. 
For every \textit{forward diffusion} step, the ground truth 3D pose \( h_0 \) is used to generate \( N \) sets of samples \(\{h_1, \dots, h_K\}\) by adding noise iteratively for \( K \) steps. This process is formulated as: 
\begin{equation}
h_k =\mu + \sqrt{\alpha _k}(h_0- \mu) + \sqrt{(1-\alpha _k)} \cdot \epsilon,
\label{eq:forward}
\end{equation}
where \(h_k\) is the $k$-th iteration sample from the noisy distribution $H_K$. \(\mu=\sum_{m=1}^M \pi_m \mu_m\) represents the mean of the GMM distribution, where \( M \) is the number of Gaussian components, \( \pi_m \) and \( \mu_m \) are the weight and mean of the \( m \)-th component respectively. \(\epsilon\) is a noise variable sampled from GMM distribution. 
$\alpha _{1:K} \in (0,1]^K$ is a fixed decreasing sequence that controls the noise scaling at each diffusion step, and \(K\) denotes the maximum number of diffusion steps. Notably, as $\alpha_K \approx 0$, ${h}_K$ converges to a sample drawn from the fitted GMM model. 
Additionally, to extract contextual dependencies from the 2D pose, we leverage a pre-trained HPE model~\cite{zhang2022mixste}, denoted as \( f_{2D} \).
Moreover, to enable the diffusion model to learn to effectively denoise samples at each diffusion step, we generate a unique step embedding $k$ for the $k$-th step using a sinusoidal function. 
The entire framework is supervised using a Mean Squared Error (MSE) loss to iteratively reconstruct \( h_{k-1} \) from \( h_k \) at each step. The diffusion loss \(\mathcal{L}_{\text{diff}}\) is defined as:  
\begin{equation}
\label{L_diff}
\mathcal{L}_{\text{diff}}  =   \sum_{k=1}^{K} \norm{\mathcal{D}_\theta ({h}_{k},f_{2D}, k) - {h}_{k-1}}_2,
\end{equation}
where \( \mathcal{D}_\theta(\cdot) \) denotes the denoising model. 

\textbf{Inference Process.}
During the inference phase, we follow a procedure similar to the initialization step in training. First, the noisy pose distribution \( H_K \) is initialized, and the 2D contextual features \( f_{2D} \) are extracted. 
Subsequently, the \textit{reverse diffusion} process is performed. Specifically, \( N \) samples are drawn from \( H_K \), denoted as \(\{h^1_K, h^2_K, \dots, h^N_K\}\). These samples are then iteratively processed by the denoising model \( \mathcal{D}_\theta \) for \( K \) steps to generate \( N \) high-quality 3D pose predictions, represented as \(\{h^1_0, h^2_0, \dots, h^N_0\}\). 
Finally, the mean of the \( N \) denoised samples \(\{h^1_0, h^2_0, \dots, h^N_0\}\) is computed to produce the final 3D pose.

\subsection{Autoregressive Pose Conditional Diffusion}
Human motion follows a continuous trajectory, with each pose influencing the subsequent ones. Consequently, maintaining consistency with previously predicted 3D poses is crucial in 3D HPE. To this end, we propose an AutoPCD model (Fig.~\ref{fig:framework}) that strategically integrates previously established 2D pose conditions and predicted 3D poses into the denoising process for enhanced accuracy and consistency.

Specifically, AutoPCD leverages both 2D pose features and the 3D pose predicted at the preceding time step as conditioning inputs for the current denoising operation. This autoregressive framework ensures that each predicted pose builds upon its predecessors, effectively capturing the sequential dependencies inherent to human motion. Acknowledging that past 2D conditions and prior 3D predictions contain valuable contextual information for subsequent frames, we introduce the HPIM to amalgamate these features. HPIM consolidates historical 2D keypoints and previous time-step predictions of 3D poses by extracting spatial-temporal features through a combination of Graph Networks and Attention Integration Networks. The integrated historical pose embedding is denoted as $f_{his}$, serving as autoregressive inputs for AutoPCD and enhancing its capacity to model temporal continuity effectively. Thus, the denoising process of our AutoPCD can be formatted as:
\begin{equation}
    h_{k-1} = \mathcal{D}_\theta(h_k, f_\text{2D}, f_{his}, k),
\end{equation}
For the detail of historical pose embedding $f_{his}$, please refer to the \textbf{Historical Pose Integration Module} section.

Furthermore, to enhance pose accuracy and physical plausibility, we propose STPG. This guidance leverages the temporal characteristics and inherent structure of the human skeleton to impose constraints, guiding the approximation of poses generated at each step of the denoising process.
Therefore, our StarPose harnesses the power of autoregressive modeling to capture the continuity of human motion, while leveraging historical information and physical constraints to improve both accuracy and coherence in 3D HPE. 

\begin{figure}
    \centering
\centerline{\includegraphics[width=\linewidth]{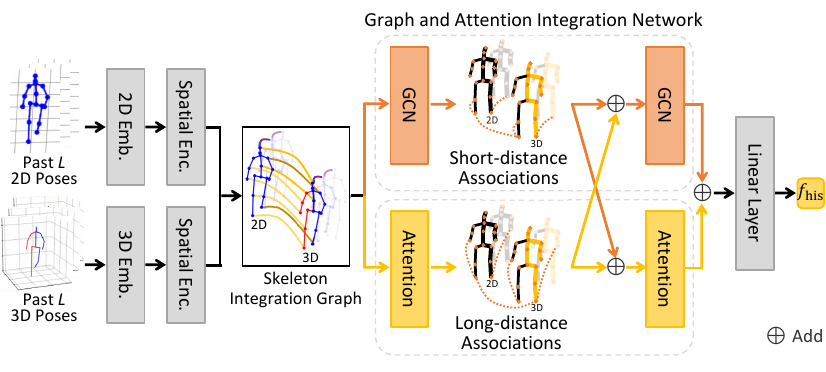}}
    \caption{Illustration of HPIM. We employ two spatial encoders to extract the features of ground-truth 2D and predicted 3D poses from past moments, respectively. By establishing the spatial-temporal relationships of these poses through a Skeleton Integration Graph, we then utilize a GCN layer and an attention layer to capture both short-distance and long-distance associations among joints. Finally, we fuse these two types of relationships to create comprehensive historical pose features. 
    }
    \label{fig:hpim}
\end{figure}

\textbf{Historical Pose Integration Module.}
\label{sec:hpim}
To maintain consistency across the predicted 3D pose sequence, we introduce the HPIM, which is designed to capture spatial and temporal dependencies among past predictions. This module models the relationships between a sequence of previously generated 3D poses and the corresponding ground-truth 2D pose sequence. By effectively integrating information from the predicted 3D history with ground-truth 2D observations, HPIM mitigates the accumulation of errors that often arise in autoregressive prediction settings. The fused representation serves as the conditioning input for the next-frame 3D pose prediction, allowing the model to autoregressively refine its estimates over time. Through this mechanism, our approach preserves temporal continuity and improves the overall coherence of the generated pose sequences.
As illustrated in Fig.~\ref{fig:hpim}, the HPIM consists of three main components:
\textit{a) Multi-dimensional Spatial Encoders}, which encode the spatial features of 2D and 3D joints in previous time steps.
\textit{b) Skeleton Integration Graph}, which fuses 2D and 3D features to enhance feature representation.
\textit{c) Graph and Attention Integration Network}, which enables bidirectional communication between spatial and temporal relationships in the human skeleton.

\textit{a) Multi-dimensional Spatial Encoders. }
For a target 3D pose \(I_t\), we utilize the past $L$ 2D poses \(P_{2D} = \{I_\tau\}_{\tau = t-L}^{t}\) and the corresponding $L$ 3D predictions \(P_{3D} = \{I_\tau\}_{\tau = t-L}^{t}\) as inputs to the HPIM, where $t$ is the current pose index and $L$ is the length of past pose sequence. \textit{Note that for the first $L$ frames of the video, we use all previous 2D poses and 3D predictions since they are less than $L$ frames. These inputs encapsulate the network's predictions over the preceding $L$ time steps, providing temporal context for \(I_t\). } 
For each frame in \(P_{2D}\), we first apply a linear projection to map the 2D joint positions into a higher-dimensional space. To encode joint positional information, we add a learnable spatial positional embedding, resulting in \(P_{2D}^\prime \in \mathbb{R}^{J \times C}\), where \(C\) represents the embedding's channel dimension. Similarly, for each frame in \(P_{3D}\), the 3D skeletons are projected into \(P_{3D}^\prime \in \mathbb{R}^{J \times C}\).  
Next, we use two spatial encoders to extract high-dimensional features independently for \(P_{2D}^\prime \) and \(P_{3D}^\prime \).   
For the past $L$ frames, the 2D spatial transformer and 3D spatial transformer produce feature sequences \(Z_{2D} \in \mathbb{R}^{L \times (J \cdot C)}\) and \(Z_{3D} \in \mathbb{R}^{L \times (J \cdot C)}\), respectively, representing the 2D and 3D joint features over time. To preserve temporal information, we then augment \(Z_{2D}\) and \(Z_{3D}\) with learnable temporal positional embeddings, ensuring that the positional context of each frame is retained.

\textit{b) Skeleton Integration Graph.} 
After encoding the topological relationships between joints, we represent the human skeleton as a graph to model the historical joint features, where each joint serves as a node in this graph. Our approach integrates the encoded outputs from both the 2D and 3D spatial encoders, constructing a unified graph that incorporates both 2D and 3D pose information.  
We represent the 2D and 3D pose feature sequences (\(Z_{2D}\) and \(Z_{3D}\)) as a skeleton integration graph \(\mathcal{G} = \{\upsilon_t, \varepsilon_t \,|\, t = 1, \dots, L\}\), where \(\upsilon_t = \{\upsilon_{t,i} \,|\, i = 1, \dots, 2J\}\) denotes the nodes (pose joints) and \(\varepsilon_t\) represents the edges that capture spatial and temporal connections.  
For spatial relationships within a single frame, we include direct connections between neighboring joints to capture the skeletal topology. Additionally, we introduce cross-dimensional edges that connect each joint in the 2D pose to its corresponding joint in the 3D pose. This enables the model to effectively learn the cross-dimensional relationships between 2D and 3D representations.  
To model motion dynamics, we add temporal edges that link the same joint across consecutive frames. These temporal connections are crucial for capturing the evolution of poses over time, allowing the graph to represent temporal changes in human motion.  
By incorporating both spatial and temporal relationships, the skeleton integration graph \(\mathcal{G}\) models dynamic human poses across dimensions, providing a comprehensive framework for learning from both 2D and 3D pose sequences.  

\textit{c) Graph and Attention Integration Network. }
We employ the Skeleton Integration Graph to establish connections across three key relationships: adjacent joints within the same dimension, corresponding joints across different dimensions, and the same joint at different time steps. We refer to these connections as \textit{short-distance associations} and use Graph Convolutional Networks (GCNs) to effectively model these localized relationships.  
However, human body movements also exhibit \textit{long-distance associations}, which involve relationships between distant joints within the same dimension, different joints across dimensions, different joints across time, and etc. Capturing these long-distance dependencies is challenging for traditional GCNs due to their localized receptive fields.  
To address this limitation, we integrate an attention mechanism to model long-distance associations, enabling the network to effectively capture complex feature relationships that span large spatial and temporal distances. 
Specifically, the network consists of two parallel branches: one employing GCNs to process the Skeleton Integration Graph \(\mathcal{G}\) and the other utilizing attention layers to learn the pose representations using \(Z\), where \(Z\) is the combined output of \(Z_{2D}\) and \(Z_{3D}\). 
Let \(f_{G}\) denote the output feature from the GCN layer and \(f_{A}\) denote the output feature from the attention layer. 
To enhance the pose representation, we combine the short-distance information from \(f_{G}\) with the long-distance information from \(f_{A}\), which can be expressed as:
\begin{equation}
    f^\prime= {f}_{G} + {f}_{A}.
\end{equation}
This integrated feature \(f'\) is then fed into the next GCN and attention layers. The integration of complementary information enhances the modeling capabilities of both the GCN and attention layers, contributing to more accurate spatial-temporal relationship learning in subsequent layers.
Finally, the outputs from the last GCN and attention layers, denoted as \(f_G^\prime\) and \(f_A^\prime\), respectively, are added together, which can be expressed as: 
\begin{equation}
    f_{his}= \text{FC}({f}_{G}^\prime + {f}_{A}^\prime),
\end{equation}
where \(\mathrm{FC}(\cdot)\) is a linear layer to scale the dimensions.  
As a result, \(f_{his}\) serves as the historical pose integration condition and is concatenated with other conditions as the input of the denoising model $\mathcal{D}_\theta$. Consequently, Eq.~\eqref{L_diff} is updated as follows:
\begin{equation}
\label{L_diff_modi}
\mathcal{L}_{\text{diff}}  =   \sum_{k=1}^{K} 
\norm{\mathcal{D}_\theta ({h}_{k},f_{2D}, f_{his}, k) - {h}_{k-1}}_2 .
\end{equation}

\subsection{Spatial-Temporal Physical Guidance}
After the AutoPCD network is established, we enhance the training process by going beyond the original denoising loss $\mathcal{L}_\text{diff}$ of the diffusion model. We introduce a novel STPG mechanism, leveraging both the inherent properties of diffusion models and the physical characteristics of 3D HPE. This guidance not only improves the training effectiveness but also plays a critical role during inference by steering the generation of more accurate 3D HPE.
Specifically, we employ an energy function~\cite{lecun2006} to quantify the discrepancy between the generated pose and the given conditions. By leveraging the gradient of this energy function, we iteratively refine the 3D pose estimation to achieve higher accuracy.

\textbf{Energy Function.} Recent studies~\cite{Yu_2023_ICCV,energyNEURIPS2022} have introduced energy functions \(\mathcal{E}\) as a means to evaluate the alignment between noisy data \(h_k\) and specified conditions \(\mathbf{c}\). A lower energy value signifies a closer alignment, with the ideal energy value being zero. Thus, the energy function acts as a metric to quantify the compatibility between the conditions \(\mathbf{c}\) and the data \(h_k\). 
Moreover, the noisy data \(h_k\) can be iteratively refined through a process known as energy guidance, which utilizes the gradient of the energy function \(\nabla_{h_k}\mathcal{E}(\mathbf{c},h_k)\)~\cite{Yu_2023_ICCV}. This refinement can be mathematically expressed as: 
\begin{equation}
    h_{k-1}=h_{k-1} - \rho_k\nabla_{h_k}\mathcal{E}(\mathbf{c}, h_k),
    \label{eq:energy_guided_sampling}
\end{equation}
where $\rho_k$ is a scale factor, which can be seen as the learning rate of the correction term. In practical scenarios, directly measuring the distance between the noisy intermediate data \(h_k\) and the conditions \(\mathbf{c}\) can be challenging. Instead, a more feasible approach is to measure the compatibility using the clean data \({h}_{0|k}\), which is derived at each step. From Eq.~\eqref{eq:forward}, the clean pose \({h}_{0|k}\) can be estimated from \(h_k\) as: 
\begin{equation}
    h_{0|k}=\mu + \frac{1}{\sqrt{{\alpha}_k}}(h_k -\mu -\sqrt{(1-{\alpha}_k)} \mathcal{D}_\theta(h_k, f_{2D}, f_{his}, k) ),
    \label{eq:x0t}
\end{equation}
Using this formulation, the compatibility between the clean pose and the conditions can be represented by the energy function \(\mathcal{E}(\mathbf{c}, h_{0|k})\).
Substituting this into the energy guidance process, the approximated reverse process can be expressed as:
\begin{equation}
    h_{k-1}=h_{k-1} - \rho_k\nabla_{h_k}\mathcal{E}(\mathbf{c}, h_{0|k}).
    \label{eq:approximated_energy_guided_sampling}
\end{equation}

\begin{figure}[t]
\centerline{\includegraphics[width=\columnwidth]{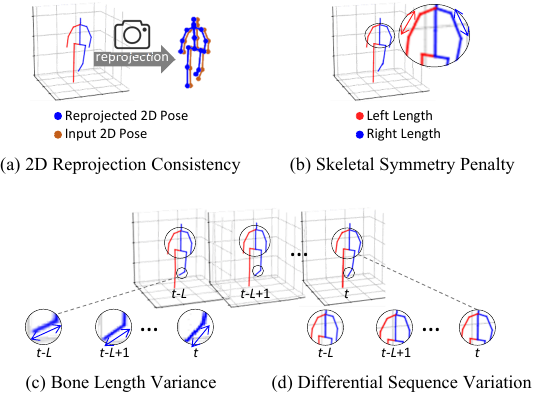}} 
    \caption{Four physical constraints of Spatio-Temporal Physical Guidance. (a) 2D Reprojection Consistency rectifies unreasonable predicted positions by projecting 3D joints into the 2D space. (b) Skeletal Symmetry Penalty renders the predicted lengths of left and right limbs more rational. (c) Bone Length Variance maintains the consistency of bone length during movement. (d) Differential Sequence Variation sustains the continuity and timing of joint movements. }
    \label{fig:stpg}
\end{figure}

In the context of 3D HPE, the stringent physical constraints governing human motion can be effectively utilized as an energy function within diffusion-based 3D HPE to refine predictions.
Specifically, we propose STPG, which leverages prior knowledge of human skeletal topology and is independent of ground-truth data. 
This includes prior such as 2D reprojected positions and skeletal symmetry. Additionally, since 3D poses are estimated across a video sequence, it is essential to ensure the temporal consistency of the human skeleton. To achieve this, we incorporate constraints based on bone length consistency and differential sequence variation. These four physical constraints are detailed in Fig.~\ref{fig:stpg} and described as follows.

\textbf{2D Reprojected Consistency.} 
The input 2D keypoints can serve as geometric priors to guide the model in generating the most likely 3D pose. Although 2D keypoints do not contain direct depth information, they provide valuable clues about the potential locations of human joints in 3D space~\cite{Zhu_2023_ICCV, Cho_2021_ICCV}. Specifically, the 3D joint should lie along the ray extending from the camera's optical center to the 2D keypoint in the image plane. Therefore, establishing the correspondence between 3D predictions and 2D observations is crucial for effectively guiding the denoising process. 
To leverage this information, we use the known intrinsic camera parameters to reproject 3D hypotheses $h_{0|k}$ onto the 2D camera plane, resulting in the reprojected 2D positions \(\mathcal{R}(h_{0|k})\). The 2D reprojected consistency error, denoted as \(\mathcal{L}_p\), is computed as the Euclidean distance between the reprojected 2D positions \(\mathcal{R}(h_{0|k})\) and the corresponding input 2D keypoints \(\bm{x}\), which can be formulated as:
\begin{equation}
       \mathcal{L}_p =\sum_{k=1}^{K}\sum_{j=1}^{J}||\mathcal{R}(h_{0|k}^{(j)})-\bm{x}^{(j)}||_{2},
    \label{loss:Lp}
\end{equation}
where $j$ denotes the joint index, and $\mathcal{R}(\cdot)$ being the reprojection function proposed by~\cite{Cho_2021_ICCV} to perform perspective transformation. 

\textbf{Skeleton Symmetry Penalty.} 
Due to the lack of depth information, 2D poses cannot reflect the lengths of bones. 
However, in 3D HPE, the symmetry of bone lengths between corresponding left and right limb pairs serves as an important prior, guiding the model to resolve ambiguities inherent in monocular estimation.
To this end, we propose a penalty term based on skeleton symmetry to guide the model in generating anatomical accuracy 3D poses. 
We begin by calculating the bone lengths of the left and right limb pairs using Euclidean distance. Specifically, we predefine $P$ sets of corresponding left and right bone pairs (represented as keypoint pairs) based on the dataset annotations.
For $p$-th bone pair, the bone length between adjacent joints $j_1$ and $j_2$ of each side \(B_p\) is calculated as:
\begin{equation}
       B_p(h_{0|k}) =||h_{0|k}^{(j_1)}-h_{0|k}^{(j_2)}||_{2}.
    \label{bone}
\end{equation}
To enforce skeleton symmetry, we define the skeleton symmetry loss \(\mathcal{L}_s\), which penalizes the differences in the bone lengths of corresponding left and right limb pairs. Mathematically, \(\mathcal{L}_s\) can be expressed as: 
\begin{equation}
       \mathcal{L}_s =\sum_{k=1}^{K}\sum_{p=1}^{P}||\text{Left}(B_{p}(h_{0|k}))-\text{Right}(B_{p}(h_{0|k}))||_{2},
    \label{loss:Ls}
\end{equation}
where \(\text{Left}(B_{p}(h_{0|k}))\) means the bone length for a left-side bone in pose \(h_{0|k}\), and \(\text{Right}(B_{p}(h_{0|k}))\) is the corresponding right-side bone length.

\textbf{Bone Length Variance.} 
Human bones exhibit consistent lengths due to skeletal structure, even during motion. 
In the context of 3D pose estimation from video sequences, maintaining temporal consistency in bone lengths is essential for generating continuous and anatomically plausible poses.
To this end, we propose a bone length variance loss, which minimizes the variation of bone lengths across frames in a sequence, thereby promoting kinematic consistency. 
To begin, we compute the bone lengths for predefined skeletal connections. Specifically, we define $Q$ sets of adjacent bone connections based on the dataset annotations, where \(Q = J-1\). For each $q$-th pair of keypoints in a frame, the bone length \(B_q\) is calculated as described in Eq.~\eqref{bone}.
Next, for a target 3D pose \(I_t\), we calculate the variance \(V(B_q)\) of each bone length within a sliding window consisting of the past $L$ poses, denoted as \(\{I_\tau\}_{\tau = t-L}^{t}\). This variance for the $q$-th bone can be formulated as:
\begin{equation}
       V(B_q) =\frac{1}{L} \sum_{l=1}^{L} || B_q(h_{0|k}^{l}) - \overline{B_q^L} ||_{2},
    \label{Vbone}
\end{equation}
where $h_{0|k}^{l}$ represents the hypothetical pose in the $l$-th frame of the sliding window, and \(\overline{B_q^L}\) is the mean bone length of $q$-th bone across $L$ frames. The final bone length variance loss \(\mathcal{L}_b\) is the cumulative variance across all $Q$ bones, given by: 
\begin{equation}
       \mathcal{L}_b =\sum_{k=1}^{K} \sum_{q=1}^Q  V(B_q).
    \label{loss:Lb}
\end{equation}
By minimizing \(\mathcal{L}_b\), the model learns to produce pose sequences with consistent bone proportions, aligning with real-world kinematic constraints.

\textbf{Differential Sequence Variation.} 
In addition to ensuring consistent bone lengths over time, different joints in the human body exhibit varying degrees of motion during dynamic activities.
To better maintain the temporal consistency of the pose sequence, we introduce dynamic constraints on joint motion in 3D HPE~\cite{hossain2018exploiting}. Specifically, we introduce a differential sequence variation loss, which guides the denoising process by weightedly calculating the 3D joint differences between adjacent frames. This loss encourages smoother motion transitions across frames and helps maintain temporal coherence in the pose sequence. The differential sequence variation loss \(\mathcal{L}_d\) can be formulated as: 
\begin{equation}
       \mathcal{L}_d =\sum_{k=1}^{K} \sum_{j=1}^{J} w_j (h_{0|k}^{t(j)} - h_{0|k}^{t-1(j)})^2.
    \label{loss:Ld}
\end{equation}
where $h_{0|k}^{t}$ and $h_{0|k}^{t-1}$ represent the hypothetical poses at the current and previous time respectively, \(w_j\) is a pre-defined weight~\cite{hossain2018exploiting} assigned to each keypoint \(j\), reflecting the varying influence of each joint's movement. 

\input{table/algo1}

In summary, the energy function we propose for 3D HPE using AutoPCD incorporates four distinct loss components, formulated as follows:  
\begin{equation}
    \mathcal{E}(\mathbf{c}, h_{0|k}) = \lambda_p \mathcal{L}_p + \lambda_s \mathcal{L}_s
    + \lambda_b \mathcal{L}_b + \lambda_d \mathcal{L}_d,
    \label{loss:e}
\end{equation}  
where \(\lambda_p\), \(\lambda_s\), \(\lambda_b\), and \(\lambda_d\) are the respective weights assigned to each loss term. The detailed algorithm is presented in \textbf{Algo.}~\ref{algo1}. This energy function enables the model to produce poses at each step that align with the desired spatial-temporal physical guidance.  

To train our StarPose framework, we define the overall loss function as:  
\begin{equation}
    \mathcal{L} = \mathcal{L}_\text{diff} + \lambda_p \mathcal{L}_p + \lambda_s \mathcal{L}_s
    + \lambda_b \mathcal{L}_b + \lambda_d \mathcal{L}_d,
    \label{loss_L}
\end{equation}  
where \(\mathcal{L}_\text{diff}\) represents the original diffusion denoising loss, and the remaining terms correspond to the individual components of the energy function.  

A notable feature of our proposed STPG is that it is fully \textbf{\textit{plug-and-play}}, which enables seamless integration into any diffusion-based 3D HPE framework. This module enhances the accuracy of 3D pose generation by leveraging the energy function \(\mathcal{E}\), which does not rely on any ground-truth data and can be directly applied during the inference phase (i.e., the reverse diffusion process). Please refer to the experiments at \textit{Section IV-E} for more details.

%% file: table/params.tex
\begin{table}[t]
\small
\setstretch{1.1}
\caption{Common Symbols and Descriptions.}
\label{para}
\centering
\resizebox{\linewidth}{!}{
\begin{tabular}{c|l}
    \hline
    \noalign{\smallskip}
    Symbols & Descriptions \\ 
    \noalign{\smallskip}
    \hline
    \noalign{\smallskip}
    $N$ & Number of hypotheses (i.e. sample numbers). \\
    $K$ & Number of iterations (i.e. diffusion step numbers). \\
    $J$ & Number of joints. \\
    $L$ & Length of past pose sequences. \\
    $t$ & Current time pose index. \\
    $\mathcal{D}_\theta$& Denosing model. \\
    $H_K$ & Initialized noise pose distribution. \\
    $h_k$ & The $k$-th iteration sample from the $H_K$.\\
    $h_{0|k}$ & Clean hypothetical pose estimated from $h_k$. \\ 
    \noalign{\smallskip}
    \hline
\end{tabular}}
\end{table}

%% file: table/algo1.tex
\begin{algorithm}[t]
\small
\caption{Reverse Process with our proposed STPG}
\label{algo1}
\setstretch{1.2}
\begin{algorithmic}[1] 
    \Require condition $\mathbf{c}$, energy function \(\mathcal{E}(\mathbf{c}, \cdot)\), denoising model $\mathcal{D}(\cdot)$, GMM distribution \(H_K\), GMM distribution mean $\mu$, noise scalaers ${\alpha}_k$, and learning rate $\rho_k$.
    \State $h_K\sim H_K$\
    \For{$k = K, ..., 1$}
        \State $h_{0|k}=\mu + \frac{1}{\sqrt{{\alpha}_k}}(h_k -\mu -\sqrt{(1-{\alpha}_k)} \mathcal{D}(h_k, f_{2D}, f_{ST}, k) )$
        \State $h_{k-1} = \sqrt{{\alpha}_{k-1}} h_{0|k} + \sqrt{1 - {\alpha}_{k-1}}  \mathcal{D}(h_k, f_{2D}, f_{ST}, k) $
        \State $\mathcal{E}(\mathbf{c}, h_{0|k}) = \lambda_p \mathcal{L}_p + \lambda_s \mathcal{L}_s
       + \lambda_b \mathcal{L}_b + \lambda_d \mathcal{L}_d$
        \State $\boldsymbol{g}_k = \nabla_{h_k}\mathcal{E}(\mathbf{c}, h_{0|k})$
        \State $h_{k-1} = h_{k-1} - \rho_k \boldsymbol{g}_k$
    \EndFor
    \State \textbf{return} $h_{0}$
\end{algorithmic}
\end{algorithm}

%% file: sec/exp.tex
\section{Experiments}
\subsection{Datasets and Evaluation Metrics}
We evaluate our method on two widely-used 3D HPE benchmarks, including Human3.6M~\cite{ionescu2013human3} and MPI-INF-3DHP~\cite{mehta2017monocular}. 

\textbf{Human3.6M}~\cite{ionescu2013human3} is a widely used benchmark for 3D HPE, comprising 3.6 million images of 11 subjects performing 15 daily activities. Following standard protocol\cite{Xu_2024_CVPR, Peng_2024_CVPR, eccv_repose, Gong_2023_CVPR, D3DP2023_ICCV, Tang_2023_CVPR}, we train on five subjects (S1, S5, S6, S7, S8) and evaluate on two unseen subjects (S9, S11). Input 2D poses are either predicted by CPN~\cite{chen2018cascaded} or taken from the ground truth.
Evaluation is based on the Mean Per Joint Position Error (MPJPE), the average Euclidean distance (in \textit{mm}) between predicted and ground-truth 3D joint coordinates, and the Procrustes-aligned MPJPE (P-MPJPE), which applies a rigid alignment before error computation.
To assess temporal consistency, we also report Mean Per Joint Velocity Error (MPJVE)~\cite{pavllo20193D} and Acceleration Error (ACC-ERR)~\cite{Mehraban_2024_WACV}, capturing smoothness through joint-wise velocity and acceleration deviations.

\textbf{MPI-INF-3DHP}~\cite{mehta2017monocular} is a more challenging dataset than Human3.6M, as it includes both indoor and outdoor scenes with diverse themes and activities. For experiments on MPI-INF-3DHP, we use ground-truth 2D poses to enable a direct comparison with prior works~\cite{Xu_2024_CVPR, Gong_2023_CVPR, D3DP2023_ICCV, Tang_2023_CVPR, GLA-GCN_2023_ICCV, tang2023ftcm}. We evaluate the performance using MPJPE, Percentage of Correct Keypoints (PCK) within a 150\( mm \) threshold, and the Area Under Curve (AUC) for a range of PCK thresholds. 

\subsection{Implementation Details}
\textbf{Architecture.} 
Following~\cite{Gong_2023_CVPR}, for the denoising model $\mathcal{D}$, we adopt GraFormer~\cite{Zhao_2022_CVPR}, a lightweight GCN-based architecture, as the backbone for performing 3D HPE via diffusion. The denoising model $\mathcal{D}$ consists of three stacked GCN-Attention Blocks, where each GCN-Attention Block includes two standard GCN layers and a Self-Attention layer. 
For the Context Encoder, we leverage a pre-trained transformer-based network MixSTE~\cite{zhang2022mixste} to capture the spatial-temporal context information $f_{2D}$ in the 2D pose sequence. The Context Encoder is pre-trained on the training set and remains frozen during the diffusion model training. 

\input{table/h36m_cpn}

\textbf{Parameter Settings.} 
The input sequence length of the 2D pose is set to \(f = 243\) and the number of joints is \(J=17\). 
We use \( N = 5 \) pose samples and \( K = 50 \) reverse diffusion steps. 
To accelerate the diffusion inference procedure, we employ the Denoising Diffusion Implicit Model (DDIM)~\cite{song2021denoising}, which reduces the required reverse diffusion steps to $5$. 
During the forward diffusion process, the decreasing sequence \(\alpha_{1:K}\) is generated using the formula: 
$\alpha_k = \prod ^{k}_{i=1}(1-{\beta_i}),$
where \(\beta_{1:K}\) is a sequence linearly interpolated between \(1e{-4}\) and \(2e{-3}\). The number of GMM components is set to $M=5$.
In the Historical Pose Integration Module and Bone Length Variance, the length of past pose sequences is set to $L=27$. 
For the hyperparameters in Eq.~\ref{loss:e} and Eq.~\ref{loss_L}, we empirically set \(\lambda_p = \lambda_s = 1\) and \(\lambda_b = \lambda_d = 0.01\). 
The denoising model \( g \) is trained for $50$ epochs using the Adam optimizer~\cite{kingma2017adam, yang2024g} with an initial learning rate of \(1e{-4}\). An exponential learning rate decay schedule with a decay factor of $0.9$ is applied after every $10$ epoch. The training is conducted on an Ubuntu 22.04 system equipped with 256 GB of RAM and an NVIDIA GeForce RTX 4090 GPU. A batch size of 1024 is utilized during training.

\input{table/pose_smooth}

\subsection{Comparisons with State-of-the-art Methods}
\textbf{Results on Human3.6M.} 
We quantitatively compare the performance of our proposed StarPose with the state-of-the-art (SOTA) methods for 3D HPE on Human3.6M~\cite{ionescu2013human3}, as shown in Table~\ref{h36m_cpn}, where the 2D poses obtained by CPN~\cite{chen2018cascaded} are used as inputs.
Our method achieves the best results 29.9$mm$ in MPJPE and 24.6$mm$ in P-MPJPE. It significantly outperforms the previous SOTA method FinePose~\cite{Xu_2024_CVPR} by 2.0$mm$ under MPJPE. 
Moreover, StarPose demonstrates significant performance improvements over transformer-based methods such as RePose~\cite{eccv_repose}, MoAGFormer~\cite{Mehraban_2024_WACV}, and STCFormer~\cite{Tang_2023_CVPR}. These results highlight the effectiveness of our diffusion model in resolving the inherent ambiguities of monocular pose estimation by generating multiple plausible 3D pose hypotheses. 
When compared with other diffusion-based approaches, including FinePose~\cite{Xu_2024_CVPR}, KTPFormer~\cite{Peng_2024_CVPR}, D3DP~\cite{D3DP2023_ICCV}, and DiffPose~\cite{Gong_2023_CVPR}, StarPose achieves superior accuracy. This improvement can be attributed to the benefits of our designed Autoregressive Pose Conditional Diffusion, which enhances the network's ability to generate accurate and continuous 3D poses through historical features and physical constraints.  
Additionally, we conduct experiments using ground truth 2D pose as input, with the results presented in Table~\ref{h36m_gt}. 
Under this setting, StarPose achieves 15.5$mm$ in MPJPE, surpassing all existing methods and demonstrating the superior capability of our framework in lifting 2D poses to 3D. These results validate the efficacy of StarPose in improving the accuracy and reliability of monocular 3D HPE.

We also evaluate temporal smoothness of 3D HPE using MPJVE~\cite{pavllo20193D} and ACC-ERR~\cite{Mehraban_2024_WACV}, which measure joint-wise velocity and acceleration deviations, respectively. As shown in Table~\ref{pose_smooth}, our method achieves notable gains of 2.6$mm/s$ and 3.7$mm/s^2$ on MPJVE and ACC-ERR, indicating superior temporal consistency. This improvement stems from the autoregressive design of StarPose and the spatiotemporal topological guidance provided by STPG, which together enforce both temporal continuity and kinematic plausibility across the predicted 3D pose sequence.

\input{table/h36m_gt}
\input{table/3dhp}

\textbf{Results on MPI-INF-3DHP.} 
We further evaluate the performance of our method on the MPI-INF-3DHP~\cite{mehta2017monocular}. As shown in Table~\ref{3dhp}, our StarPose achieves the best results with PCK of 98.9\%, AUC of 82.5\%, MPJPE of 20.8$mm$, which significantly outperforms the previous SOTA method FinePose~\cite{Xu_2024_CVPR} by 2.3$mm$ under MPJPE. 
These results are consistent with the performance trends observed on the Human3.6M, further validating the strong generalization capability of our proposed method across different datasets.

\subsection{Qualitative Results}
\begin{figure}[!t]
\centerline{\includegraphics[width=\columnwidth]{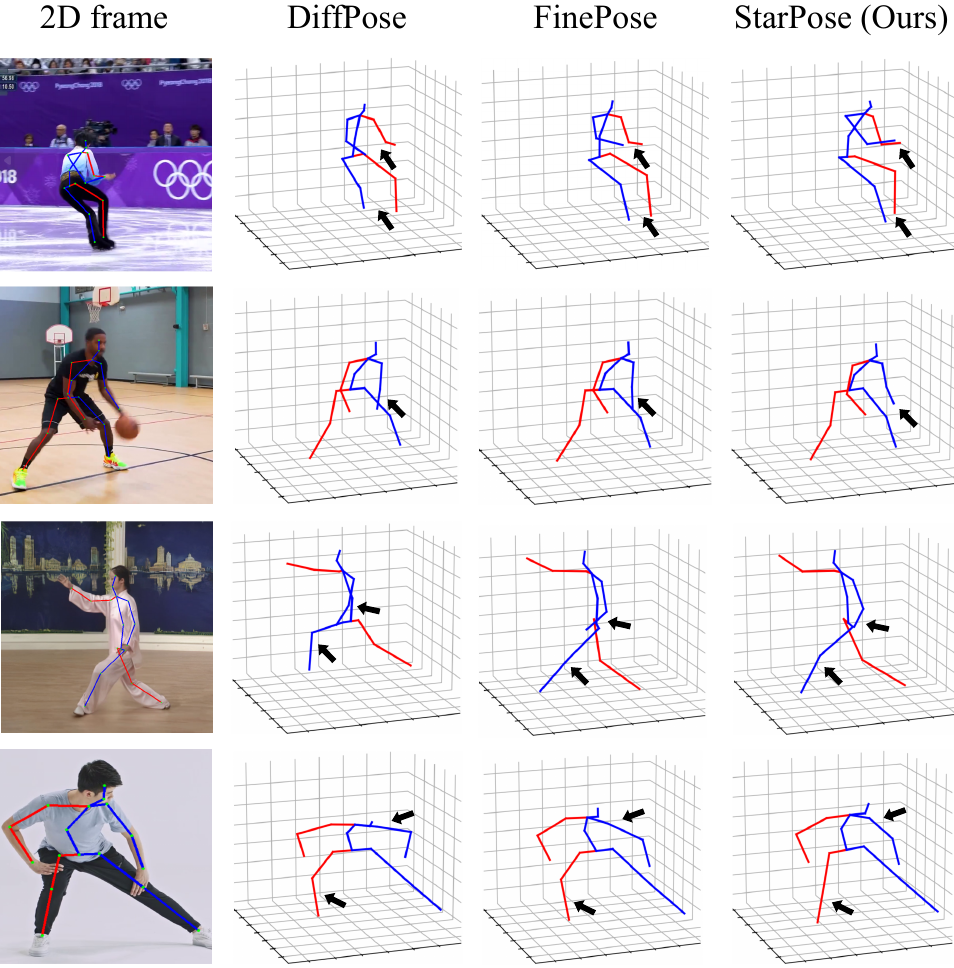}} 
    \caption{Qualitative comparisons among DiffPose~\cite{Gong_2023_CVPR}, FinePose~\cite{Xu_2024_CVPR}, and our StarPose on in-the-wild videos across diverse activities. The black arrows indicate regions where our method yields visibly superior results.}
    \label{wild}
\end{figure}

To further assess the generalization capability of our model, we conduct 3D HPE on real-world, in-the-wild video sequences downloaded from the Internet. These sequences feature a wide range of sports scenes characterized by extensive body movements, intricate postures, and rapid dynamics.
As shown in Fig.~\ref{wild}, our method is evaluated against leading state-of-the-art approaches, including DiffPose\cite{Gong_2023_CVPR} and FinePose~\cite{Xu_2024_CVPR}. The visual comparisons clearly demonstrate that our approach achieves more precise localization of joints and more faithfully preserves the spatial relationships among joints over time.
Importantly, StarPose consistently surpasses existing methods under these challenging conditions, underscoring the effectiveness of AutoPCD in capturing enriched spatio-temporal features. In addition, the integration of STPG plays a vital role in preserving anatomically plausible joint configurations and ensuring temporal consistency, which is essential for generating stable and realistic motion trajectories.
These findings collectively demonstrate the robustness and adaptability of our framework in tackling complex real-world scenarios.

\subsection{Plug-and-Play of Spatial-Temporal Physical Guidance} 
Our proposed STPG is designed as a \textit{plug-and-play} component, as it does not require additional training. Note that STPG is fully compatible with the multiple iterative denoising steps inherent in the diffusion model, allowing the pose in each subsequent iteration to be optimized by calculating the gradient of the energy function. This allows it to be seamlessly integrated into any diffusion-based 3D HPE framework to enhance the accuracy of 3D pose generation. 
To evaluate its effectiveness, we conduct experiments by adding the STPG module to the diffusion-based 3D pose estimation methods DiffPose~\cite{Gong_2023_CVPR} and D3DP~\cite{D3DP2023_ICCV}. The results, presented in Table~\ref{plugandplay}, show that the MPJPE errors of both methods were significantly reduced by incorporating our STPG, indicating its effectiveness in guiding the denoising process. Since STPG does not modify the underlying model architecture, the integration results in only a slight reduction in frames per second (FPS). 
Notably, while the reverse process in D3DP~\cite{D3DP2023_ICCV} begins with standard Gaussian noise—unlike the initialization distributions used in ours and DiffPose~\cite{Gong_2023_CVPR}, and the added STPG achieves a significant reduction of 3.9$mm$ MPJPE on D3DP~\cite{D3DP2023_ICCV}. This highlights the effectiveness of our proposed STPG in improving the quality of 3D pose generation via human topology and dynamics constraints. 
\input{table/plugandplay}

\input{table/ablation1}
\subsection{Ablation Study}
\textbf{Effectiveness of Main Proposals.} 
We conduct an ablation study on Human3.6M~\cite{ionescu2013human3} using 2D poses from CPN~\cite{chen2018cascaded} as input ($f = 243$). Table~\ref{ablation1} evaluates the contributions of STPG and HPIM through several variants of StarPose: \textit{Baseline}, \textit{+STPG-T}, \textit{+STPG-I}, \textit{+STPG}, \textit{+HPIM}, \textit{HPIM w/o 2D}, and \textit{Pred. Intr.}.

The Baseline excludes both STPG and HPIM. Adding STPG during training (\textit{+STPG-T}) reduces the error by 0.5$mm$, confirming its regularization effect. Incorporating STPG only during inference (\textit{+STPG-I}) leads to a larger 2.9$mm$ drop in MPJPE, demonstrating its efficacy in enforcing topological consistency and suppressing implausible poses at test time. When applied jointly during training and inference (\textit{+STPG}), STPG achieves a 3.2$mm$ improvement, highlighting its full potential when used throughout the pipeline.

We then examine the efficacy of HPIM. Adding HPIM alone to the Baseline yields a substantial 6.1$mm$ gain, validating the effectiveness of its spatial-temporal historical features. By modeling autoregressive dependencies and integrating 2D-3D correlations, HPIM significantly enhances pose estimation stability and accuracy. Removing the 2D sequence from HPIM (\textit{HPIM w/o 2D}) results in degraded performance, indicating that the 2D-3D interaction is crucial for mitigating error accumulation.

To ensure a fair comparison, we use ground-truth camera intrinsics in the 2D reprojected consistency loss as D3DP~\cite{D3DP2023_ICCV}. When unavailable, we resort to a learned intrinsic prediction model~\cite{yang2023camerapose} (\textit{Pred. Intr.}). Despite a moderate performance drop, this variant still outperforms the other state-of-the-art methods, demonstrating the robustness of our framework under practical settings.

Finally, combining STPG and HPIM achieves the best performance, with a 9.6$mm$ improvement over the Baseline, which underscores the complementary contributions of STPG and HPIM.

\begin{figure}[!t]
    \centerline{\includegraphics[width=\columnwidth]{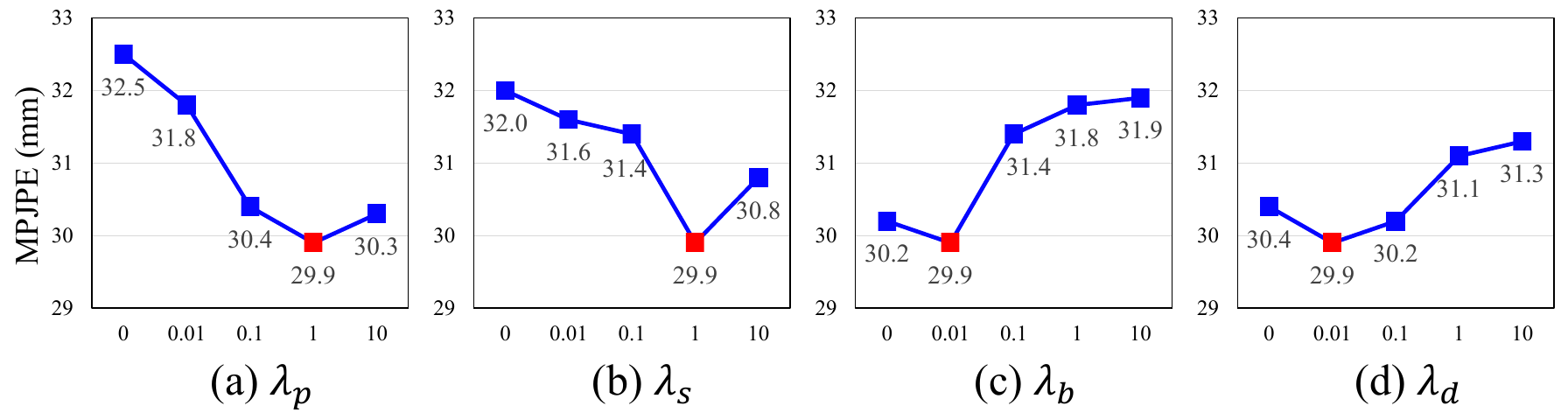}} 
        \caption{Weight sensitity analysis on (a) $\lambda_{p}$, (b) $\lambda_{s}$, (c) $\lambda_{b}$ and (d) $\lambda_{d}$ in STPG components. Red indicates the default settings.}
        \label{lambda}
    \end{figure}

\textbf{Impact of Spatial-Temporal Physical Guidance.} 
The STPG module introduces four key hyperparameters that regulate the influence of distinct loss components: $\lambda_{p}$ for 2D Reprojected Consistency, $\lambda_{s}$ for Skeleton Symmetry Penalty, $\lambda_{b}$ for Bone Length Variance, and $\lambda_{d}$ for Differential Sequence Variation. To assess their individual impact, we perform ablation studies by varying one hyperparameter at a time, while fixing the others to their default values ($\lambda_{p} = 1$, $\lambda_{s} = 1$, $\lambda_{b} = 0.01$, $\lambda_{d} = 0.01$).
The results, presented in Fig.~\ref{lambda}, show that each component contributes meaningfully to the overall performance when appropriately weighted. Notably, the default configuration, indicated by red markers, yields the highest accuracy. This suggests that balancing the loss terms on a comparable scale enables the model to integrate structural constraints and temporal dynamics effectively, without favoring any single objective.

These findings highlight the complementary roles of the loss components. Overemphasizing one term, such as temporal consistency or anatomical regularity, can compromise other aspects of the learning process, leading to degraded generalization. In contrast, our balanced design enables STPG to jointly enforce spatial validity, motion smoothness, and kinematic consistency, resulting in improved accuracy and robustness across diverse motion patterns.

\begin{figure}[!t]
    \centerline{\includegraphics[width=\columnwidth]{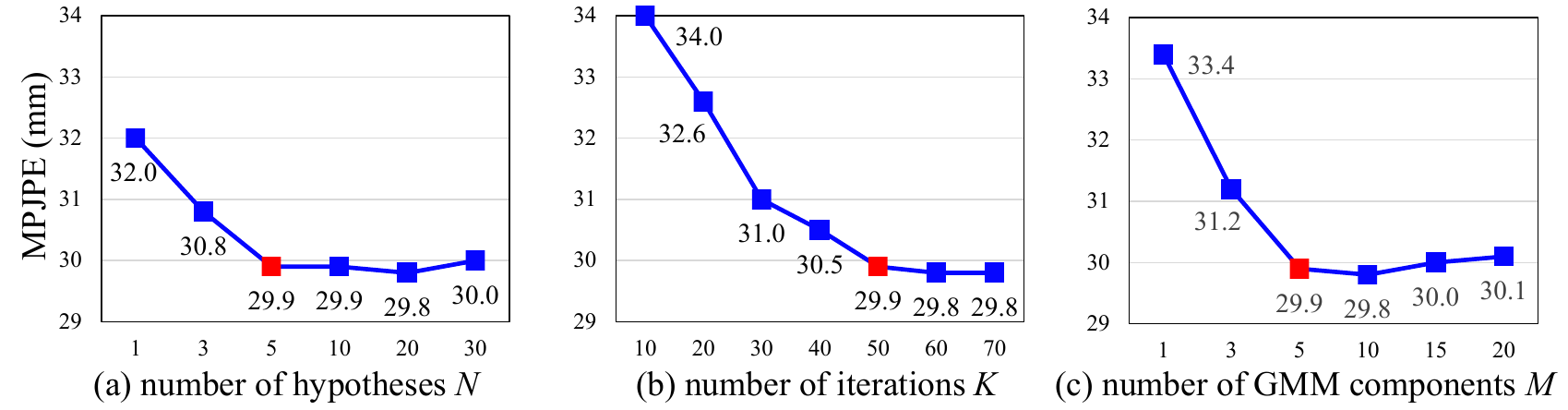}} 
        \caption{Ablations on the (a) number of hypotheses $N$, (b) the numbers of iterations $K$, and (c) the number of GMM components $M$. Red indicates the default settings. }
        \label{NandK}
    \end{figure}

\textbf{Number of hypotheses $N$, iterations $K$ and GMM components $M$.} 
Fig.~\ref{NandK}(a) illustrates how performance varies with the number of hypotheses $N$ (i.e., sample numbers). Performance improves consistently as $N$ increases, reaching a plateau at $N = 5$. Accordingly, we fix $N = 5$ to ensure diverse candidate poses while maintaining computational efficiency.
Fig.~\ref{NandK}(b) presents the impact of the number of iterations $K$ (i.e., diffusion steps). MPJPE decreases sharply up to $K = 50$, beyond which gains become marginal. This indicates convergence of the refinement process. We therefore set $K = 50$, balancing accuracy and efficiency in resolving pose ambiguity.
Fig.~\ref{NandK}(c) shows the effect of the number of GMM components $M$. Increasing $M$ from 1 to 5 steadily reduces MPJPE, reflecting improved modeling of the pose distribution. However, further increasing $M$ to 10 yields negligible benefits. We thus choose $M = 5$ as it sufficiently captures the underlying distribution without unnecessary complexity.

\subsection{Complexity Analysis} 

\input{table/ablation5}

Table~\ref{ablation5} presents a comparison of the inference speed of StarPose with existing methods, measured in Frames Per Second (FPS). All experiments were conducted on a single NVIDIA GeForce RTX 4090 GPU, and the DDIM technique~\cite{song2021denoising} was employed. 
Our StarPose achieves an impressive inference speed of 1370 FPS, which is competitive with existing methods. 
When compared to transformer-based approaches~\cite{li2022mhformer, shan2022p}, StarPose not only outperforms them in terms of accuracy but also maintains competitive speed. In comparison with the diffusion-based method D3DP~\cite{D3DP2023_ICCV}, StarPose achieves superior accuracy while significantly accelerating the inference process, meeting the real-time requirements of most applications. Although our method incurs a slight decrease in FPS relative to DiffPose~\cite{Gong_2023_CVPR}, it results in a substantial improvement in accuracy, with a reduction of 7.0$mm$ in MPJPE.
Additionally, we evaluated the impact of each individual module on the inference speed, as shown in the last column of Table~\ref{ablation1}. The inclusion of the STPG module results in a minor decrease of 116 FPS, as the STPG does not involve model inference but rather computes the distance between the predicted pose and the given conditions. The HPIM also causes a small reduction in FPS, but this impact is minimal due to its lightweight design. These results demonstrate that our method achieves a significant accuracy improvement while retaining the ability to perform real-time pose estimation.

\subsection{Limitations}

\begin{figure}[!t]
    \centerline{\includegraphics[width=\columnwidth]{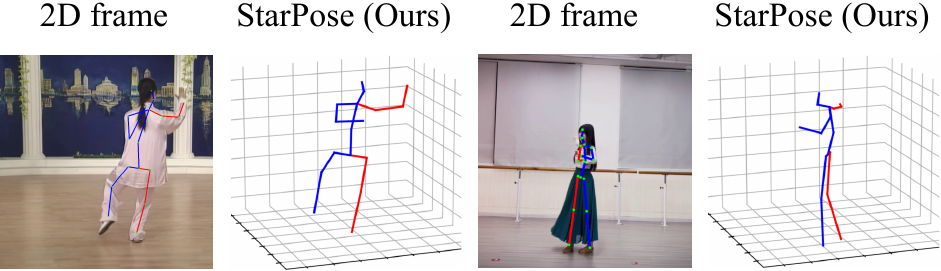}} 
        \caption{Failure case analysis. Our method may fail under severe motion occlusions, where inaccuracies in 2D pose estimations lead to erroneous 3D reconstructions.}
        \label{faircase}
    \end{figure}

Although StarPose exhibits strong performance in 3D HPE, certain limitations remain. In particular, severe motion occlusion can significantly degrade the quality of the 2D pose estimations produced by the upstream detector, which in turn leads to inaccuracies in the corresponding 3D reconstructions, as illustrated in Fig.~\ref{faircase}. Addressing these challenges remains a complex and open research problem. We plan to investigate more robust strategies for 3D pose estimation under such challenging conditions in future work.

%% file: table/h36m_cpn.tex
\begin{table*}[t]
\centering
\setstretch{1.1}
\caption{Comparisons with Previous Methods on Human3.6M. The 2D poses obtained by CPN~\cite{chen2018cascaded} are used as inputs.  Top: Using MPJPE as a metric.  Bottom: Using P-MPJPE as a metric.  \(f\): The input 2D sequence's length.  (*): Diffusion-based methods. 
 \textcolor{red}{Red}: Best.  \textcolor{blue}{Blue}: Second best.}
\vspace{-20pt}
\begin{center}
\resizebox{\textwidth}{!}{
\begin{tabular}{lcc|ccccccccccccccc|c}
\hline
\noalign{\smallskip}
{\textbf{MPJPE}} & Reference &$f$& Dir. & Disc. & Eat & Greet & Phone & Photo & Pose & Pur. & Sit & SitD. & Smoke & Wait & WalkD. & Walk & WalkT. & Avg$\downarrow$ \\
\noalign{\smallskip}
\hline
\noalign{\smallskip}
PoseFormer~\cite{zheng20213D} & ICCV'21 &81&
41.5&44.8&39.8&42.5&46.5&51.6&42.1&42.0&53.3&60.7&45.5&43.3&46.1&31.8&32.2&44.3\\
RIE~\cite{shan2021improving} &MM'21 &243&
40.8&44.5&41.4&42.7&46.3&55.6&41.8&41.9&53.7&60.8&45.0&41.5&44.8&30.8&31.9&44.3\\
Anatomy~\cite{chen2021anatomy} &TCSVT'21 &243&
41.4&43.5&40.1&42.9&46.6&51.9&41.7&42.3&53.9&60.2&45.4&41.7&46.0&31.5&32.7&44.1\\
STE~\cite{li2022exploiting} &TMM'22 &351&
39.9&43.4&40.0&40.9&46.4&50.6&42.1&39.8&55.8&61.6&44.9&43.3&44.9&29.9&30.3&43.6\\
U-CDGCN~\cite{hu2021conditional} &MM'21 &96&
38.0&43.3&39.1&39.4&45.8&53.6&41.4&41.4&55.5&61.9&44.6&41.9&44.5&31.6&29.4&43.4\\
MHFormer~\cite{li2022mhformer} &CVPR'22 &351&
39.2&43.1&40.1&40.9&44.9&51.2&40.6&41.3&53.5&60.3&43.7&41.1&43.8&29.8&30.6&43.0\\
FTCM~\cite{tang2023ftcm} &TCSVT'24 & 351&
39.0&41.1&38.6&41.6&45.6&50.4&41.4&37.8&52.9&65.4&42.4&41.2&42.9&31.7&32.9&43.0\\
P-STMO~\cite{shan2022p} &ECCV'22 &243&
38.9&42.7&40.4&41.1&45.6&49.7&40.9&39.9&55.5&59.4&44.9&42.2&42.7&29.4&29.4&42.8\\
HDFormer~\cite{ijcai23hdformer}&IJCAI'23&96&
38.1&43.1&39.3&39.4&44.3&49.1&41.3&40.8&53.1&62.1&43.3&41.8&43.1&31.0&29.7&42.6\\
MixSTE~\cite{zhang2022mixste}&CVPR'22 &243&
37.6&40.9&37.3&39.7&42.3&49.9&40.1&39.8&51.7&55.0&42.1&39.8&41.0&27.9&27.9&40.9\\
DUE~\cite{zhang2022uncertainty} &MM'22 &300&
37.9&41.9&36.8&39.5&40.8&49.2&40.1&40.7&47.9&53.3&40.2&41.1&40.3&30.8&28.6&40.6\\
STCFormer~\cite{Tang_2023_CVPR} & CVPR'23 & 243&
39.6&41.6&37.4&38.8&43.1&51.1&39.1&39.7&51.4&57.4&41.8&38.5&40.7&27.1&28.6&41.0\\
D3DP*~\cite{D3DP2023_ICCV} &ICCV'23&243&
37.3&39.5&35.6&37.8&41.3&48.2&39.1&37.6&49.9&52.8&41.2&39.2&39.4&27.2&27.1&39.5\\
MoAGFormer~\cite{Mehraban_2024_WACV} &WACV'24 &243&
36.8&38.5&35.9&33.0&41.1&48.6&38.0&34.8&49.0&51.4&40.3&37.4&36.3&27.2&27.2&38.4\\
DiffPose*~\cite{Gong_2023_CVPR} & CVPR'23 & 243&
33.2&36.6&33.0&35.6&37.6&45.1&35.7&35.5&46.4&49.9&37.3&35.6&36.5&24.4&24.1&36.9\\
RePose~\cite{eccv_repose} & ECCV'24 & 243&
34.6&36.8&35.0&31.3&38.8&44.7&35.3&35.5&47.5&50.9&38.4&35.5&34.9&23.9&25.0&36.5\\
KTPFormer*~\cite{Peng_2024_CVPR} & CVPR'24 & 243&
\textcolor{blue}{30.1}&32.1&29.1&30.6&35.4&39.3&32.8&30.9&43.1&45.5&34.7&33.2&32.7&\textcolor{blue}{22.1}&23.0&33.0\\
FinePose*~\cite{Xu_2024_CVPR} & CVPR'24 & 243&
31.4&\textcolor{blue}{31.5}&\textcolor{blue}{28.8}&\textcolor{blue}{29.7}&\textcolor{blue}{34.3}&\textcolor{blue}{36.5}&\textcolor{blue}{29.2}&\textcolor{blue}{30.0}&\textcolor{blue}{42.0}&\textcolor{blue}{42.5}&\textcolor{blue}{33.3}&\textcolor{blue}{31.9}&\textcolor{blue}{31.4}&22.6&\textcolor{blue}{22.7}&\textcolor{blue}{31.9}\\

\bestcell{\textbf{StarPose}}&\bestcell{\textbf{Ours}}&\bestcell{243}&
\bestcell{\textcolor{red}{28.7}}&\bestcell{\textcolor{red}{28.4}}&\bestcell{\textcolor{red}{27.2}}&\bestcell{\textcolor{red}{27.1}}&\bestcell{\textcolor{red}{32.1}}&\bestcell{\textcolor{red}{35.1}}&\bestcell{\textcolor{red}{28.9}}&\bestcell{\textcolor{red}{29.7}}&\bestcell{\textcolor{red}{39.4}}&\bestcell{\textcolor{red}{40.6}}&\bestcell{\textcolor{red}{30.4}}&\bestcell{\textcolor{red}{29.5}}&\bestcell{\textcolor{red}{29.7}}&\bestcell{\textcolor{red}{20.9}}&\bestcell{\textcolor{red}{21.4}}&\bestcell{\textcolor{red}{29.9}}
\\
\noalign{\smallskip}
\hline
\noalign{\smallskip}
\noalign{\smallskip}

\hline
\noalign{\smallskip}
{\textbf{P-MPJPE}} & Reference &$f$& Dir. & Disc. & Eat & Greet & Phone & Photo & Pose & Pur. & Sit & SitD. & Smoke & Wait & WalkD. & Walk & WalkT. & Avg$\downarrow$ \\
\noalign{\smallskip}
\hline
\noalign{\smallskip}
STE~\cite{li2022exploiting} &TMM'22 &351&
32.7&35.5&32.5&35.4&35.9&41.6&33.0&31.9&45.1&50.1&36.3&33.5&35.1&23.9&25.0&35.2\\
FTCM~\cite{tang2023ftcm} &TCSVT'24 & 351&
31.9&35.0&34.0&34.2&35.9&42.1&32.3&31.2&46.5&51.8&36.4&33.7&34.3&23.9&24.9&35.2\\
RIE~\cite{shan2021improving} &MM'21 &243&
32.5&36.2&33.2&35.3&35.6&42.1&32.6&31.9&42.6&47.9&36.6&32.1&34.8&24.2&25.8&35.0\\
Anatomy~\cite{chen2021anatomy} &TCSVT'21 &243&
32.6&35.1&32.8&35.4&36.3&40.4&32.4&32.3&42.7&49.0&36.8&32.4&36.0&24.9&26.5&35.0\\
PoseFormer~\cite{zheng20213D} & ICCV'21 &81&
32.5&34.8&32.6&34.6&35.3&39.5&32.1&32.0&42.8&48.5&34.8&32.4&35.3&24.5&36.0&34.6\\
MHFormer~\cite{li2022mhformer} &CVPR'22 &351&
31.5&34.9&32.8&33.6&35.3&39.6&32.0&32.3&43.5&48.7&36.4&32.6&34.3&23.9&25.1&34.4\\
P-STMO~\cite{shan2022p} &ECCV'22 &243&
31.3&35.2&32.9&33.9&35.4&39.3&32.5&31.5&44.6&48.2&36.3&32.9&34.4&23.8&23.9&34.4\\
U-CDGCN~\cite{hu2021conditional} &MM'21 &96&
29.8&34.4&31.9&31.5&35.1&40.0&30.3&30.8&42.6&49.0&35.9&31.8&35.0&25.7&23.6&33.8\\
HDFormer~\cite{ijcai23hdformer}&IJCAI'23&96&
29.6&33.8&31.7&31.3&33.7&37.7&30.6&31.0&41.4&47.6&35.0&30.9&33.7&25.3&23.6&33.1\\
MixSTE~\cite{zhang2022mixste}&CVPR'22 &243&
30.8&33.1&30.3&31.8&33.1&39.1&31.1&30.5&42.5&44.5&34.0&30.8&32.7&22.1&22.9&32.6\\
DUE~\cite{zhang2022uncertainty} &MM'22 &300&
30.3&34.6&29.6&31.7&31.6&38.9&31.8&31.9&39.2&42.8&32.1&32.6&31.4&25.1&23.8&32.5\\
MoAGFormer~\cite{Mehraban_2024_WACV} &WACV'24 &243&
31.0&32.6&31.0&27.9&34.0&38.7&31.5&30.0&41.4&45.4&34.8&30.8&31.3&22.8&23.2&32.5\\
STCFormer~\cite{Tang_2023_CVPR} & CVPR'23 & 243&
29.5&33.2&30.6&31.0&33.0&38.0&30.4&29.4&41.8&45.2&33.6&29.5&31.6&21.3&22.6&32.0\\
D3DP*~\cite{D3DP2023_ICCV} &ICCV'23&243&
30.6&32.4&29.2&30.9&31.9&37.4&30.2&29.3&40.4&43.2&33.2&30.4&31.3&21.5&22.3&31.6\\
RePose~\cite{eccv_repose} & ECCV'24 & 243&
29.1&30.9&29.4&26.6&32.1&35.8&28.6&30.6&39.9&44.9&33.3&29.0&29.6&20.1&21.2&30.7\\
DiffPose*~\cite{Gong_2023_CVPR} & CVPR'23 & 243&
26.3&29.0&26.1&27.8&28.4&34.6&26.9&26.5&36.8&39.2&29.4&26.8&28.4&18.6&19.2&28.7\\
FinePose*~\cite{Xu_2024_CVPR}& CVPR'24 & 243&
27.5&28.9&26.0&27.4&29.0&34.2&27.0&26.1&35.4&37.8&29.2&26.2&25.6&19.8&20.1&28.0\\
KTPFormer*~\cite{Peng_2024_CVPR} & CVPR'24 & 243&
\textcolor{blue}{24.1}&\textcolor{blue}{26.7}&\textcolor{blue}{24.2}&\textcolor{blue}{24.9}&\textcolor{blue}{27.3}&\textcolor{blue}{30.6}&\textcolor{blue}{25.2}&\textcolor{blue}{23.4}&\textcolor{blue}{34.1}&\textcolor{blue}{35.9}&\textcolor{blue}{28.1}&\textcolor{blue}{25.3}&\textcolor{blue}{25.9}&\textcolor{blue}{17.8}&\textcolor{blue}{18.8}&\textcolor{blue}{26.2}\\
\bestcell{\textbf{StarPose}}&\bestcell{\textbf{Ours}}&\bestcell{243}&
\bestcell{\textcolor{red}{23.5}}&\bestcell{\textcolor{red}{25.7}}&\bestcell{\textcolor{red}{22.6}}&\bestcell{\textcolor{red}{23.2}}&\bestcell{\textcolor{red}{25.4}}&\bestcell{\textcolor{red}{28.7}}&\bestcell{\textcolor{red}{24.0}}&\bestcell{\textcolor{red}{23.1}}&\bestcell{\textcolor{red}{31.7}}&\bestcell{\textcolor{red}{33.2}}&\bestcell{\textcolor{red}{26.4}}&\bestcell{\textcolor{red}{24.1}}&\bestcell{\textcolor{red}{23.3}}&\bestcell{\textcolor{red}{17.3}}&\bestcell{\textcolor{red}{17.9}}&\bestcell{\textcolor{red}{24.6}}
\\
\noalign{\smallskip}
\hline

\end{tabular}}
\end{center}
\label{h36m_cpn}
\end{table*}

%% file: table/pose_smooth.tex
\begin{table*}[t]
    \centering
    \setstretch{1.1}
    \caption{Temporal Smoothness Comparisons with Previous Methods on Human3.6M. Top: Using Mean Per Joint Velocity Error (MPJVE, mm/s) as a metric. Bottom: Using Acceleration Error (ACC-ERR, mm/s$^2$) as a metric. (*): Diffusion-based methods.}
    \vspace{-20pt}
    \begin{center}
    \resizebox{\textwidth}{!}{
    \begin{tabular}{lcc|ccccccccccccccc|c}
    \hline
    \noalign{\smallskip}
    {\textbf{MPJVE}} & Reference &$f$& Dir. & Disc. & Eat & Greet & Phone & Photo & Pose & Pur. & Sit & SitD. & Smoke & Wait & WalkD. & Walk & WalkT. & Avg$\downarrow$ \\
    \noalign{\smallskip}
    \hline
    \noalign{\smallskip}
    FinePose*~\cite{Xu_2024_CVPR} & CVPR'24 & 243 & 
    5.4 & 5.6 & 5.1 & 5.9 & 4.8 & 5.8 & 6.2 & 5.9 & 5.2 & 6.6 & 4.7 & 4.5 & 6.8 & 7.6 & 6.0 & 5.7 \\
    DiffPose*~\cite{Gong_2023_CVPR} & CVPR'23 & 243 & 
    5.1 & 5.2 & 4.8 & 5.5 & 4.5 & 4.8 & 5.6 & 5.7 & 4.8 & 6.2 & 4.3 & 4.0 & 6.6 & 7.3 & 5.3 & 5.3 \\
    D3DP*~\cite{D3DP2023_ICCV} & ICCV'23 & 243 & 
    5.0 & 5.1 & 4.5 & 4.8 & 4.2 & 4.6 & 5.9 & 5.8 & 4.7 & 6.1 & 4.2 & 4.0 & 6.7 & 6.9 & 5.1 & 5.2 \\
    RePose~\cite{eccv_repose} & ECCV'24 & 243 &  
    4.5 & 4.8 & 4.0 & 4.2 & 3.5 & 3.8 & 4.3 & 5.8 & 4.1 & 5.3 & 3.7 & 3.5 & 5.5 & 5.7 & 4.7 & 4.5 \\
    KTPFormer*~\cite{Peng_2024_CVPR} & CVPR'24 & 243 & 
    \textcolor{blue}{4.1} & \textcolor{blue}{4.3} & \textcolor{blue}{3.5} & \textcolor{blue}{3.8} & \textcolor{blue}{3.3} & \textcolor{blue}{3.5} & \textcolor{blue}{3.8} & \textcolor{blue}{4.0} & \textcolor{blue}{3.9} & \textcolor{blue}{4.4} & \textcolor{blue}{3.2} & \textcolor{blue}{2.7} & \textcolor{blue}{4.9} & \textcolor{blue}{5.0} & \textcolor{blue}{4.2} & \textcolor{blue}{3.9} \\    
    \bestcell{\textbf{StarPose}} & \bestcell{\textbf{Ours}} & \bestcell{243} & 
    \bestcell{\textcolor{red}{1.4}} & \bestcell{\textcolor{red}{1.5}} & \bestcell{\textcolor{red}{1.1}} & \bestcell{\textcolor{red}{1.7}} & \bestcell{\textcolor{red}{1.2}} & \bestcell{\textcolor{red}{1.5}} & \bestcell{\textcolor{red}{1.3}} & \bestcell{\textcolor{red}{1.4}} & \bestcell{\textcolor{red}{1.2}} & \bestcell{\textcolor{red}{1.4}} & \bestcell{\textcolor{red}{1.2}} & \bestcell{\textcolor{red}{1.3}} & \bestcell{\textcolor{red}{1.7}} & \bestcell{\textcolor{red}{1.5}} & \bestcell{\textcolor{red}{1.4}} & \bestcell{\textcolor{red}{1.3}} \\
    \noalign{\smallskip}
    \hline
    \noalign{\smallskip}
    
    \hline
    \noalign{\smallskip}
    {\textbf{ACC-ERR}} & Reference &$f$& Dir. & Disc. & Eat & Greet & Phone & Photo & Pose & Pur. & Sit & SitD. & Smoke & Wait & WalkD. & Walk & WalkT. & Avg$\downarrow$ \\
    \noalign{\smallskip}
    \hline
    \noalign{\smallskip}
    FinePose*~\cite{Xu_2024_CVPR} & CVPR'24 & 243 & 
    6.0 & 6.3 & 5.8 & 6.5 & 5.8 & 6.6 & 6.9 & 6.5 & 6.1 & 7.8 & 5.7 & 5.8 & 8.0 & 7.9 & 7.0 & 6.6 \\
    DiffPose*~\cite{Gong_2023_CVPR} & CVPR'23 & 243 & 
    5.6 & 5.6 & 5.0 & 5.9 & 5.0 & 5.3 & 6.3 & 6.4 & 5.7 & 7.5 & 4.8 & 5.5 & 7.6 & 7.9 & 6.6 & 6.1 \\
    D3DP*~\cite{D3DP2023_ICCV} & ICCV'23 & 243 & 
    5.5 & 5.7 & 4.9 & 5.2 & \textcolor{blue}{4.7} & 4.9 & 6.3 & 6.3 & 5.6 & 7.0 & 4.7 & 5.0 & 7.8 & 7.8 & 6.1 & 5.8 \\
    RePose~\cite{eccv_repose} & ECCV'24 & 243 &  
    5.1 & 5.5 & 4.6 & 5.1 & 4.9 & 5.1 & 5.8 & \textcolor{blue}{6.2} & 5.3 & 6.6 & 4.5 & 4.5 & 7.5 & \textcolor{blue}{7.3} & 5.8 & 5.6 \\
    KTPFormer*~\cite{Peng_2024_CVPR} & CVPR'24 & 243 & 
    \textcolor{blue}{4.8} & \textcolor{blue}{5.2} & \textcolor{blue}{4.5} & \textcolor{blue}{4.8} & \textcolor{blue}{4.7} & \textcolor{blue}{4.7} & \textcolor{blue}{5.6} & \textcolor{blue}{6.2} & \textcolor{blue}{4.9} & \textcolor{blue}{6.2} & \textcolor{blue}{3.9} & \textcolor{blue}{4.2} & \textcolor{blue}{7.3} & 7.4 & \textcolor{blue}{5.5} & \textcolor{blue}{5.3} \\    
    \bestcell{\textbf{StarPose}} & \bestcell{\textbf{Ours}} & \bestcell{243} & 
    \bestcell{\textcolor{red}{1.7}} & \bestcell{\textcolor{red}{1.9}} & \bestcell{\textcolor{red}{1.4}} & \bestcell{\textcolor{red}{1.8}} & \bestcell{\textcolor{red}{1.5}} & \bestcell{\textcolor{red}{1.7}} & \bestcell{\textcolor{red}{1.8}} & \bestcell{\textcolor{red}{1.7}} & \bestcell{\textcolor{red}{1.3}} & \bestcell{\textcolor{red}{1.6}} & \bestcell{\textcolor{red}{1.5}} & \bestcell{\textcolor{red}{1.4}} & \bestcell{\textcolor{red}{2.3}} & \bestcell{\textcolor{red}{2.0}} & \bestcell{\textcolor{red}{1.7}} & \bestcell{\textcolor{red}{1.6}} \\
    \noalign{\smallskip}
    \hline
    \end{tabular}}
    \end{center}
    \label{pose_smooth}
\end{table*}

%% file: table/h36m_gt.tex
\begin{table}[t]
\small
\setstretch{1.1}
\setlength{\tabcolsep}{12pt}
\centering
\caption{Comparisons with Previous Methods on Human3.6M. Using ground-truth 2D poses as inputs.  \(f\): The input 2D sequence's length. (*): Diffusion-based methods.}
\begin{center}
{\begin{tabular}{lcc|c}
\hline\noalign{\smallskip}
{Method}&Reference&$f$&MPJPE$\downarrow$\\
\noalign{\smallskip}
\hline
\noalign{\smallskip}
Anatomy~\cite{chen2021anatomy} &TCSVT'21 &243&32.3\\
PoseFormer~\cite{zheng20213D} & ICCV'21 &81&31.3\\
MHFormer~\cite{li2022mhformer} &CVPR'22 &351&30.5\\
RIE~\cite{shan2021improving} &MM'21 &243&30.1\\
P-STMO~\cite{shan2022p} & ECCV'22 &243 &29.3\\
STE~\cite{li2022exploiting} &TMM'22 &351&28.5\\
FTCM~\cite{tang2023ftcm} &TCSVT'24 & 351&25.0\\
DUE~\cite{zhang2022uncertainty} &MM'22 &300&22.0\\
STCFormer~\cite{Tang_2023_CVPR} & CVPR'23 & 243&22.0\\
MixSTE~\cite{zhang2022mixste}&CVPR'22 &243&21.6\\
HDFormer~\cite{ijcai23hdformer}&IJCAI'23&96&21.6\\
D3DP*~\cite{D3DP2023_ICCV} &ICCV'23&243&19.6\\
DiffPose*~\cite{Gong_2023_CVPR} & CVPR'23 & 243&18.9\\
KTPFormer*~\cite{Peng_2024_CVPR}& CVPR'24 & 243&18.1\\
MoAGFormer~\cite{Mehraban_2024_WACV} &WACV'24 &243&17.3\\
FinePose*~\cite{Xu_2024_CVPR} & CVPR'24 & 243&16.7\\
RePose~\cite{eccv_repose} & ECCV'24 & 243&\textcolor{blue}{15.7}\\

\bestcell{\textbf{StarPose}}&\bestcell{\textbf{Ours}}&\bestcell{243}&
\bestcell{\textcolor{red}{15.5}}
\\
\noalign{\smallskip}
\hline
\end{tabular}}
\end{center}
\label{h36m_gt}
\end{table}

%% file: table/3dhp.tex
\begin{table}[t]
\small
\setstretch{1.1}
\centering
\caption{Comparisons with Previous Methods on MPI-INF-3DHP. Using ground-truth 2D poses as inputs. \(f\): the input 2D sequence's length. (*): Diffusion-based methods. }
\vspace{-20pt}
\begin{center}
\resizebox{\linewidth}{!}{\begin{tabular}{lcc|ccc}
\hline\noalign{\smallskip}
{Method}&Reference&$f$&PCK$\uparrow$&AUC$\uparrow$&MPJPE$\downarrow$\\
\noalign{\smallskip}
\hline
\noalign{\smallskip}
Anatomy~\cite{chen2021anatomy} & TCSVT'21&81&87.9&54.0&78.8\\
MixSTE~\cite{zhang2022mixste} & CVPR'22&27&94.4&66.5&54.9\\
U-CDGCN~\cite{hu2021conditional} & MM'21&96&97.9&69.5&42.5\\
HDFormer~\cite{ijcai23hdformer}&IJCAI'23&96&\textcolor{blue}{98.7}&72.9&37.2\\
P-STMO~\cite{shan2022p} & ECCV'22&81&97.9&75.8&32.2\\
FTCM~\cite{tang2023ftcm}&TCSVT'24&81&97.9&79.8&31.2\\
D3DP*~\cite{D3DP2023_ICCV}&ICCV'23&243&97.7&78.2&29.7\\
DiffPose*~\cite{Gong_2023_CVPR}&CVPR'23&81&98.0&75.9&29.1\\
GLA-GCN~\cite{GLA-GCN_2023_ICCV}&ICCV'23&81&98.5&79.1&27.7\\
FinePose*~\cite{Xu_2024_CVPR}&CVPR'24&243&\textcolor{red}{98.9}&80.0&26.2\\
STCFormer~\cite{Tang_2023_CVPR}&CVPR'23&81&\textcolor{blue}{98.7}&\textcolor{red}{83.9}&\textcolor{blue}{23.1}\\

\bestcell{\textbf{StarPose}}&\bestcell{\textbf{Ours}}&\bestcell{81}&\bestcell{\textcolor{red}{98.9}}&\bestcell{\textcolor{blue}{82.5}}&\bestcell{\textcolor{red}{20.8}}\\

\noalign{\smallskip}
\hline
\end{tabular}}
\end{center}
\label{3dhp}
\end{table}

%% file: table/plugandplay.tex
\begin{table}[t]
\small
\setstretch{1.1}
\setlength{\tabcolsep}{8.1pt}
\caption{Application of Plug-and-Play Module in Diffusion-based Methods on Human3.6M. ($\ddagger$): Our reproduction based on open source code.}
\label{plugandplay}
\centering
\begin{tabular}{lc|ll}
    \hline
    \noalign{\smallskip}
    Method & $f$ &  MPJPE$\downarrow$ & FPS$\uparrow$ \\ 
    \noalign{\smallskip}
    \hline
    \noalign{\smallskip}
    DiffPose~\cite{Gong_2023_CVPR} $\ddagger$ &243 &  39.5 & 1918\\ 
    DiffPose w/ \textcolor{blue}{STPG}   &243& 36.3 (\textcolor{blue}{$\downarrow$8.1$\%$}) & 1802 (\textcolor{blue}{$\downarrow$6.0$\%$})\\ 
    \noalign{\smallskip}
    \hline
    \noalign{\smallskip}
    D3DP~\cite{D3DP2023_ICCV} &243& 39.1 & 102\\ 
    D3DP w/ \textcolor{blue}{STPG} &243& 35.2 (\textcolor{blue}{$\downarrow$10.0$\%$}) & 98 (\textcolor{blue}{$\downarrow$3.9$\%$})\\ 
    \noalign{\smallskip}
    \hline
\end{tabular}
\end{table}

%% file: table/ablation1.tex
\begin{table}[t]
\small
\setstretch{1.1}
\setlength{\tabcolsep}{4pt}
\caption{Ablation Study of Each Component on Human3.6M.}
\label{ablation1}
\centering
\begin{tabular}{l|ccc|lc}
    \hline
    \noalign{\smallskip}
    Method & STPG-T & STPG-I  & HPIM & MPJPE$\downarrow$ & FPS$\uparrow$\\ 
    \noalign{\smallskip}
    \hline
    \noalign{\smallskip}
    Baseline  & &  &  & 39.5 & 1918\\ 
    w/ STPG-T & \checkmark & &  & 39.0    (\textcolor{blue}{-0.5}) & 1918\\ 
    w/ STPG-I&  &  \checkmark&  & 36.6  (\textcolor{blue}{-2.9}) & 1802\\ 
    w/ STPG  & \checkmark &\checkmark & & 36.3  (\textcolor{blue}{-3.2}) & 1802\\ 
    w/ HPIM  &  & & \checkmark & {33.4}  (\textcolor{blue}{-6.1}) & 1507\\ 
    HPIM w/o 2D  & \checkmark  & \checkmark & \checkmark & {31.2}  (\textcolor{blue}{-7.8}) & 1411\\ 
    Pred. Intr. & \checkmark & \checkmark& \checkmark & \textcolor{blue}{30.5}  (\textcolor{blue}{-9.0}) & 1370 \\
    Full & \checkmark & \checkmark& \checkmark & \textcolor{red}{29.9} 
 (\textcolor{blue}{-9.6}) & 1370\\ 
 \noalign{\smallskip}
    \hline
\end{tabular}
\end{table}

%% file: table/ablation5.tex
\begin{table}[!t]
\small
\setstretch{1.1}
\setlength{\tabcolsep}{1.7pt}
\caption{Ablation Study of Inference Speed in Human3.6M.}
\label{ablation5}
\centering
\begin{tabular}{lccc|cccc}
    \hline
    \noalign{\smallskip}
    Method & \(f\) & \(N\) & \(K\) &  MPJPE$\downarrow$  &Params(M)$\downarrow$ &FLOPs(G)$\downarrow$ & FPS$\uparrow$\\ 
    \noalign{\smallskip}
    \hline
    \noalign{\smallskip}
    MHFormer~\cite{li2022mhformer}  &351 &3 &-        &43.0  &24.7 &19.2 &825\\ 
    P-STMO~\cite{shan2022p} &243 &-&-            &42.8 & 6.7 & 3.5 &3960\\
    D3DP~\cite{D3DP2023_ICCV}  &243 &20 &10          &39.5  &34.6 &456.4 &102\\ 
    DiffPose~\cite{Gong_2023_CVPR} & 243 &5 & 50           & {36.9} &38.2  &17.4  & 1918  \\ 
    \textbf{Ours}&243  &5 &50          & {29.9} &49.5 &20.2 &1370 \\
    \noalign{\smallskip}
    \hline
\end{tabular}
\end{table}

%% file: sec/conclusion.tex
\section{Conclusion}
In this paper, we introduced StarPose, a novel spatial-temporal autoregressive diffusion framework designed to address the challenges of 3D HPE from monocular 2D pose inputs. By combining autoregressive modeling with spatial-temporal physical guidance, StarPose effectively captures the temporal continuity of human motion while maintaining anatomical plausibility and structural coherence.
Extensive experiments validate its superior performance, making StarPose a promising solution for accurate 3D HPE.

%% file: sec/bio.tex
\begin{IEEEbiography}[{\includegraphics[width=1in,height=1.25in,clip,keepaspectratio]{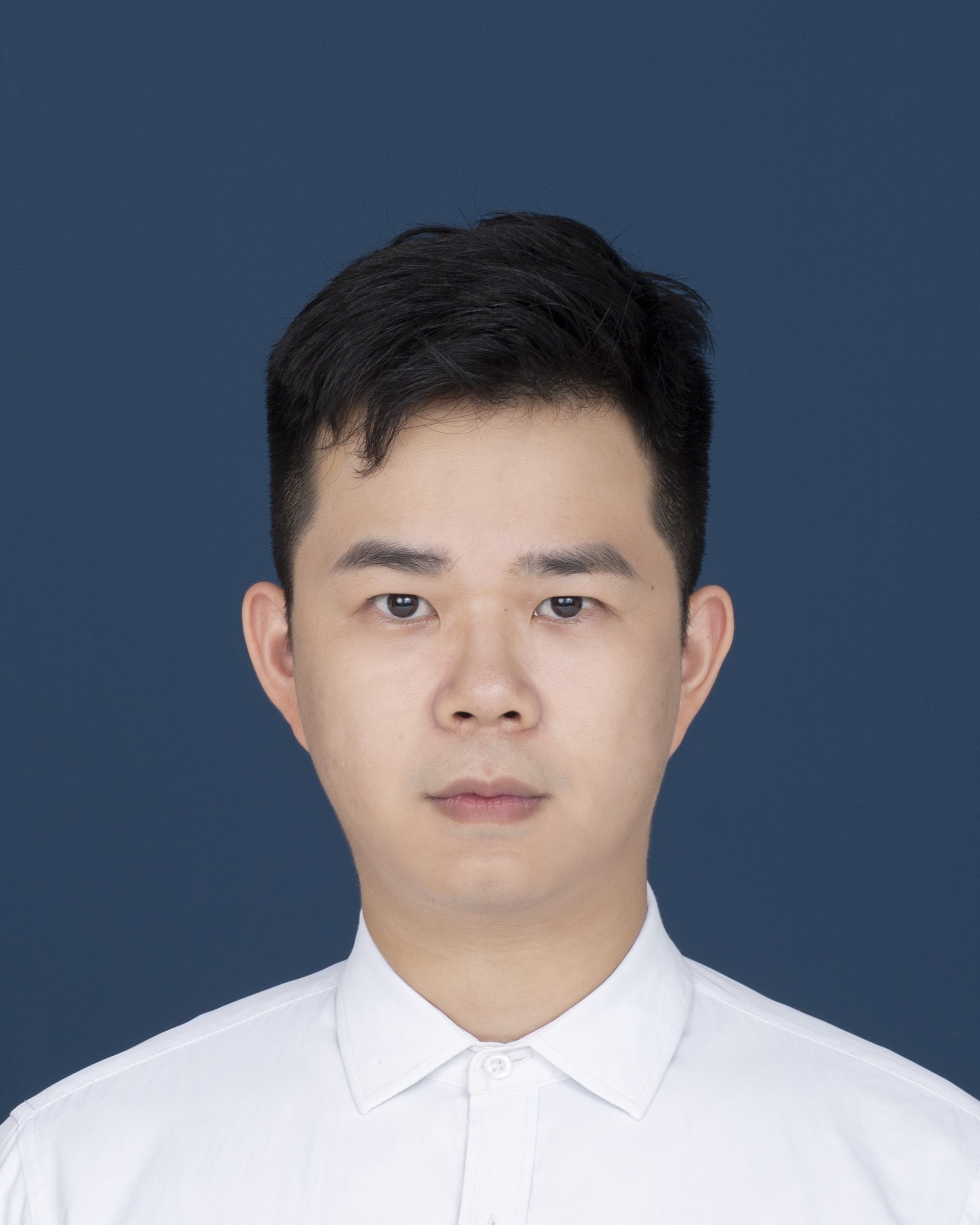}}]{Haoxin Yang} is a Ph.D. student at School of Computer Science and Engineering, South China University of Technology. He obtained his B.Sc. and M.Sc. degrees from South China Agricultural University and Shenzhen University in 2019 and 2022, respectively. His research interests include computer vision and generative models.
\end{IEEEbiography}

\begin{IEEEbiography}[{\includegraphics[width=1in,height=1.25in,clip,keepaspectratio]{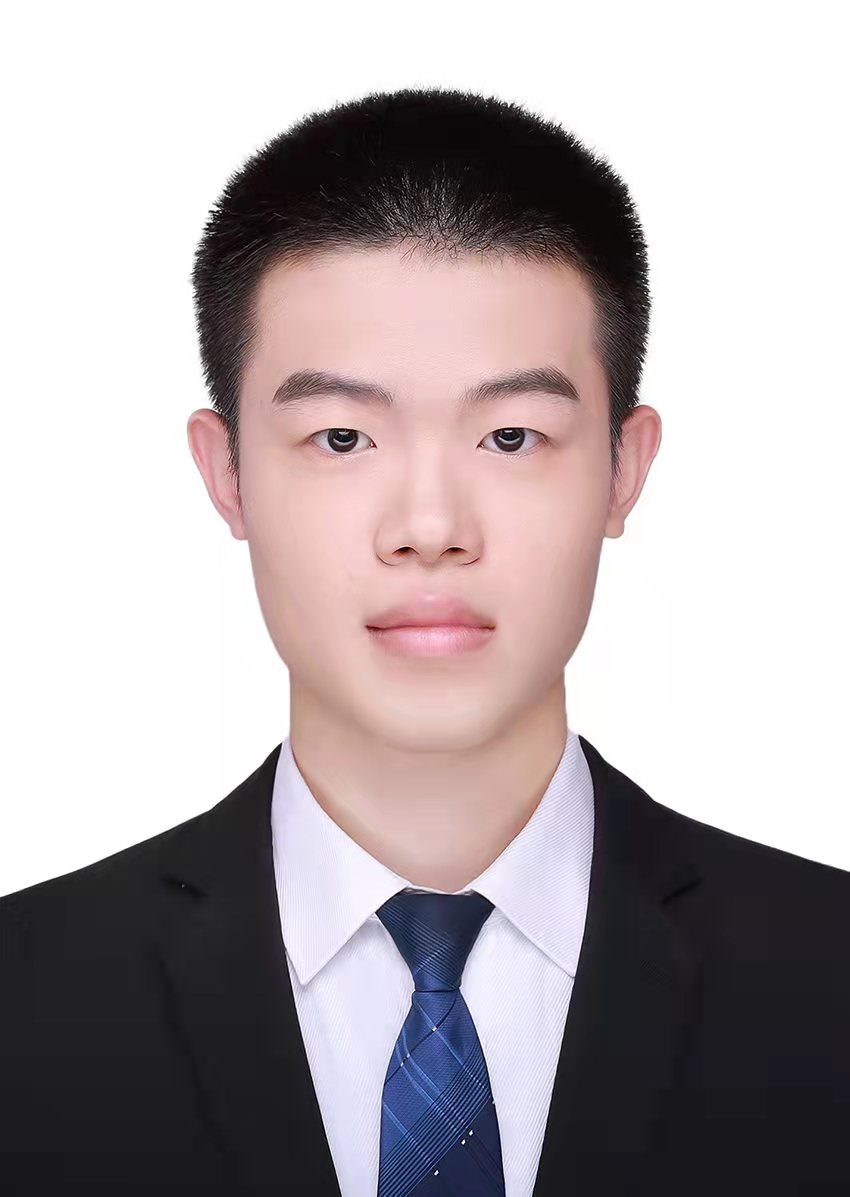}}]{Weihong Chen}
is a graduate student at the School of Computer Science and Engineering, South China University of Technology. He received his B.Sc. degree from Guangdong University of Technology in 2023. His research interests include computer vision, pose estimation, and deep learning.
\end{IEEEbiography}

\begin{IEEEbiography}[{\includegraphics[width=1in,height=1.25in,clip,keepaspectratio]{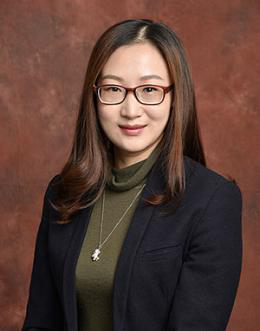}}]{Xuemiao Xu} received her B.S. and M.S. degrees in Computer Science and Engineering from South China University of Technology in 2002 and 2005 respectively, and Ph.D. degree in Computer Science and Engineering from The Chinese University of Hong Kong in 2009. She is currently a professor in the School of Computer Science and Engineering, South China University of Technology. Her research interests include object detection, tracking, recognition, and image, video understanding and synthesis, particularly their applications in the intelligent transportation.
\end{IEEEbiography}

\begin{IEEEbiography}[{\includegraphics[width=1in,height=1.25in,clip,keepaspectratio]{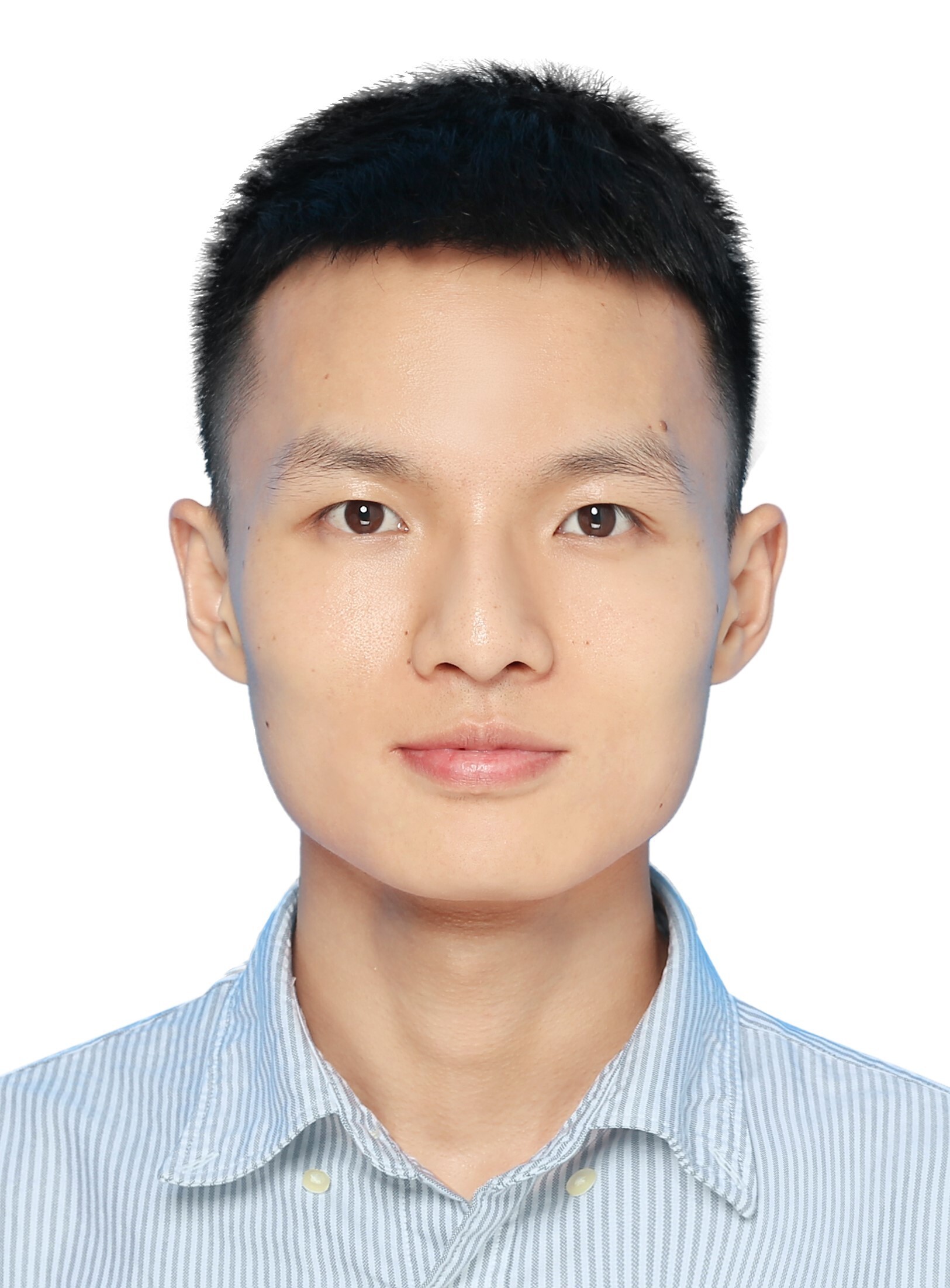}}]{Cheng Xu} received his Ph.D. degree in Computer Science and Engineering from South China University of Technology, China, in 2023. He is currently a Post-Doctoral Fellow at The Hong Kong Polytechnic University. His research interests include computer vision, image processing, and deep learning.
\end{IEEEbiography}

\begin{IEEEbiography}[{\includegraphics[width=1in,height=1.25in,clip,keepaspectratio]{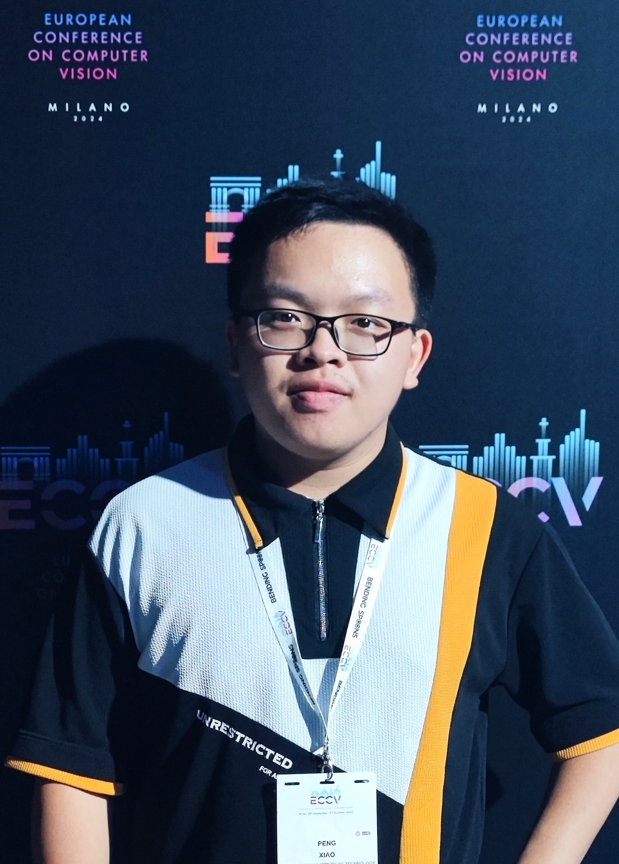}}]{Peng Xiao}
    is a graduate student at the School of Computer Science and Engineering, South China University of Technology. He received his B.Sc. degree from South China University of Technology in 2023. His research interests include computer vision, human pose, and intelligent manufacturing.
    \end{IEEEbiography}

\begin{IEEEbiography}[{\includegraphics[width=1in,height=1.25in,clip,keepaspectratio]{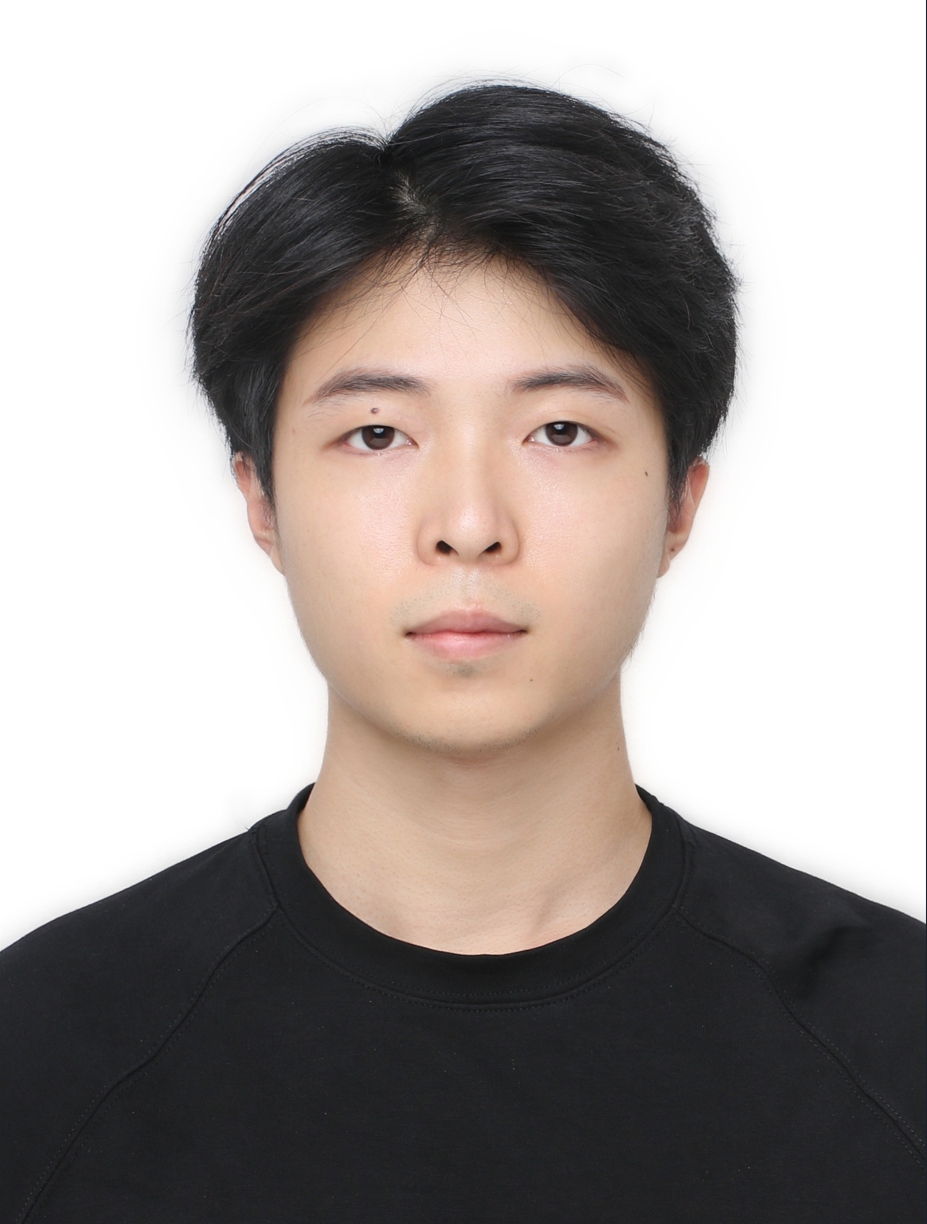}}]{Cuifeng Sun} received his B.E. degree in Software Engineering from Beijing Normal University, Zhuhai, China, in 2018. He is currently a Research Associate at Cloud Computing Center, Chinese Academy of Sciences, Dongguan, China. His research interests include artificial intelligence, large models, computer vision, image processing, and deep learning.
\end{IEEEbiography}

\begin{IEEEbiography}[{\includegraphics[width=1in,height=1.25in,clip,keepaspectratio]{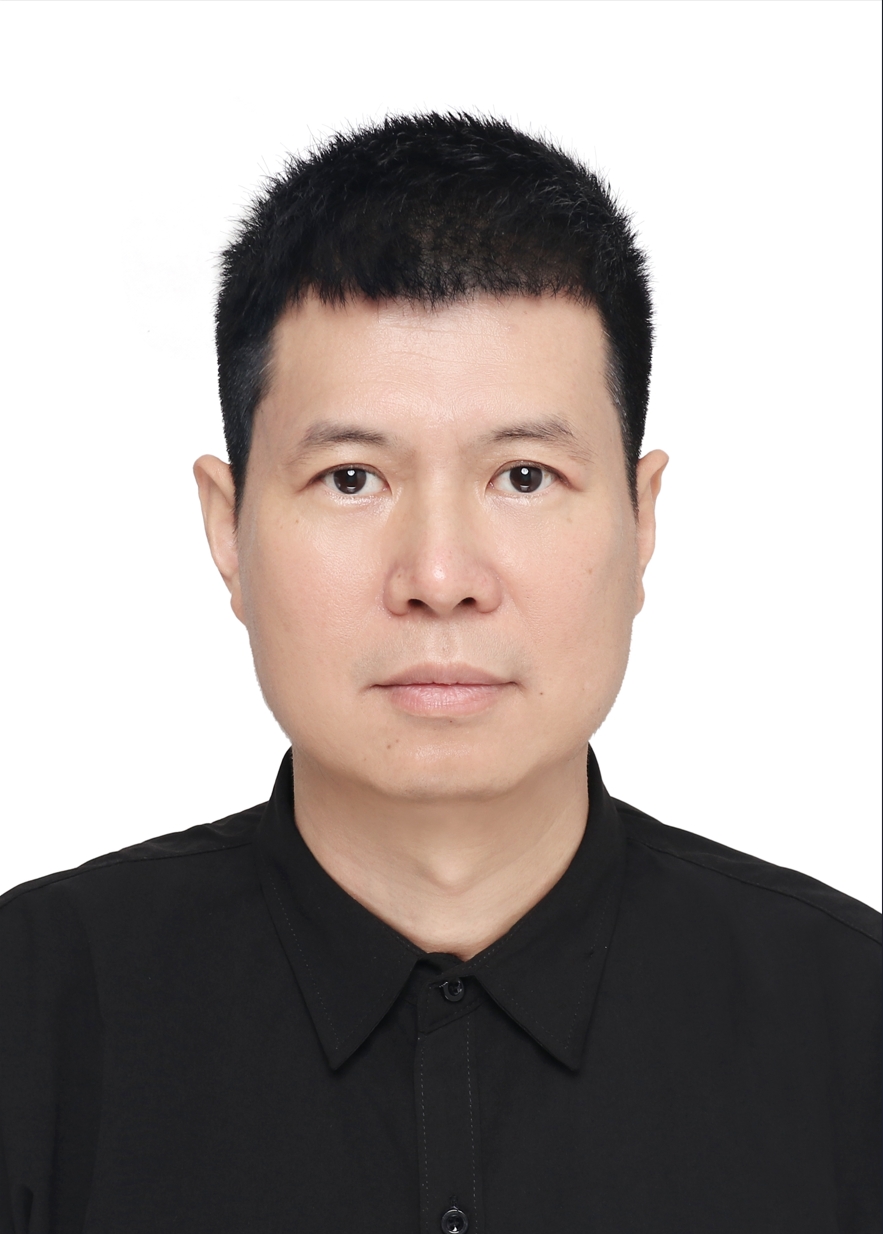}}]{Shaoyu Huang}
obtained his B.Sc. and M.Sc. degrees from South China University of Technology and Chinese Academy of Sciences Guangzhou Branch in 2000 and 2007, respectively. His research interests include object detection, tracking, recognition, and their applications in the intelligent transportation.
\end{IEEEbiography}

\begin{IEEEbiography}[{\includegraphics[width=1in,height=1.25in,clip,keepaspectratio]{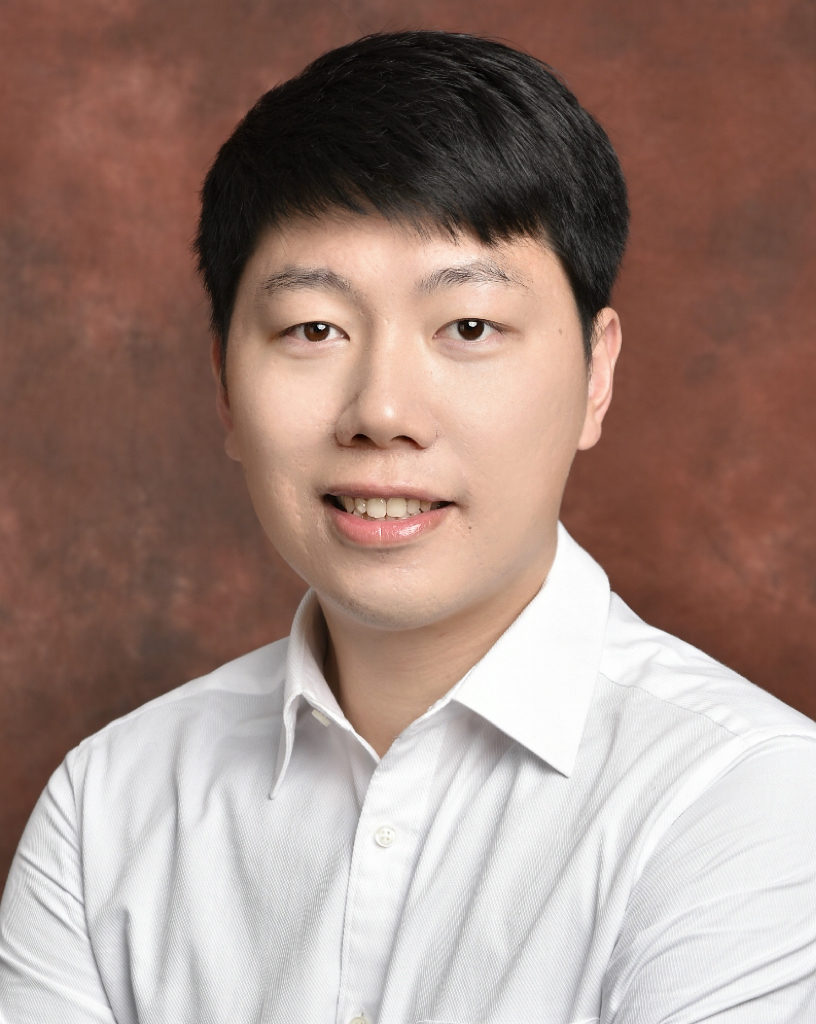}}]{Shengfeng He (Senior Member, IEEE)} is an associate professor in the School of Computing and Information Systems at Singapore Management University. Previously, he was a faculty member at South China University of Technology (2016--2022). He earned his B.Sc. and M.Sc. from Macau University of Science and Technology (2009, 2011) and a Ph.D. from City University of Hong Kong (2015). His research focuses on computer vision and generative models. He has received awards including the Google Research Award, PerCom 2024 Best Paper Award, and the Lee Kong Chian Fellowship. He is a senior IEEE member and distinguished CCF member. He serves as lead guest editor for IJCV and associate editor for IEEE TNNLS, IEEE TCSVT, Visual Intelligence, and Neurocomputing. He is an area chair/senior PC member for NeurIPS, ICML, AAAI, IJCAI, and BMVC, and will serve as Conference Chair of Pacific Graphics 2026.\end{IEEEbiography}